%% file: HCD.tex
\documentclass[conference]{IEEEtran}
\IEEEoverridecommandlockouts
\usepackage{cite}
\usepackage{amsmath,amssymb,amsfonts}
\usepackage{algorithmic}
\usepackage{graphicx}
\usepackage{textcomp}
\usepackage{xcolor}
\usepackage{setspace}
\usepackage{caption}

\def\BibTeX{{\rm B\kern-.05em{\sc i\kern-.025em b}\kern-.08em
    T\kern-.1667em\lower.7ex\hbox{E}\kern-.125emX}}

\let\labelindent\relax
\usepackage{enumitem}
    
\input{macro}

\begin{document}

\title{ Detecting Hierarchical Changes\\in Latent Variable Models  }

\author{\IEEEauthorblockN{ Shintaro Fukushima }
\IEEEauthorblockA{ \textit{Graduate School of Information Science and Technology} \\
The University of Tokyo, Tokyo, Japan \\
Email: sfukushim@gmail.com}
\and
\IEEEauthorblockN{ Kenji Yamanishi }
\IEEEauthorblockA{ \textit{Graduate School of Information Science and Technology} \\
The University of Tokyo, Tokyo, Japan \\
Email: 
yamanishi@g.ecc.u-tokyo.ac.jp }
}

\maketitle

\begin{abstract}
This paper addresses the issue of detecting hierarchical changes 
in latent variable models (HCDL) from data streams. 
There are three different levels of changes for latent variable models: 
1) the first level is the change in data distribution for fixed latent variables, 
2) the second one is that in the distribution over latent variables, 
and 3) the third one is that in the number of latent variables.
It is important to detect these changes 
because we can analyze the causes of changes 
by identifying which level a change comes from 
({\em change interpretability}). 
This paper proposes an information-theoretic framework 
for detecting changes of the three levels in a hierarchical way. 
The key idea to realize it is to employ the MDL (minimum description length) 
change statistics for measuring the degree of change, 
in combination with DNML (decomposed normalized maximum likelihood) code-length calculation. 
We give a theoretical basis for making reliable alarms for changes. 
Focusing on stochastic block models, 
we employ synthetic and benchmark datasets 
to empirically demonstrate the effectiveness of our framework 
in terms of change interpretability as well as change detection. 
\end{abstract}


\input{1_Introduction}

\input{2_Problem_Setting}

\input{3_Information_Theoretic_Methods_for_Change_Detection}

\input{4_Theory_for_Making_Reliable_Alarms}

\input{5_Hierarchical_Change_Detection}

\input{6_Experiments}

\input{7_Conclusion}

\input{8_Acknowledgement}

\bibliographystyle{IEEEtran}
\bibliography{HCD}


\end{document}

%% file: macro.tex
\usepackage{amsthm}
\theoremstyle{definition}
\newtheorem{definition}{Definition}[section]
\newtheorem{theorem}{Theorem}[section]

\newtheorem{example}{Theorem}[section]

\usepackage{amsmath,amssymb}
\usepackage{bm}

\usepackage{algorithm}
\usepackage{algorithmic}
\usepackage{graphicx}

\usepackage{mathtools}
\mathtoolsset{showonlyrefs,showmanualtags}

\usepackage{bbm}

\usepackage{multirow}
\usepackage{booktabs}

\usepackage{subcaption}

\usepackage{url}

\usepackage{lipsum}

\usepackage{textcomp}
\def\BibTeX{{\rm B\kern-.05em{\sc i\kern-.025em b}\kern-.08em
    T\kern-.1667em\lower.7ex\hbox{E}\kern-.125emX}}

\newcommand\mydef{\mathrel{\overset{\makebox[0pt]{\mbox{\normalfont\tiny\sffamily def}}}{=}}}

\newcommand{\argmin}{\operatornamewithlimits{argmin}}

\makeatletter
\def\env@cases{%
  \let\@ifnextchar\new@ifnextchar
  \left\lbrace
  \def\arraystretch{0.85}%
  \array{l@{\quad}l@{}}
}
\makeatother

%% file: 1_Introduction.tex
\section{Introduction} 

\subsection{Motivation}
\label{subsection:motivation_hcdl}

We are concerned with the issue of detecting changes in latent variable models. 
In the areas of knowledge discovery and data mining, 
latent variable models play a central role with various applications, 
such as stochastic block model (SBM)~\cite{Snijders1997} for networks, 
latent Dirichlet allocation (LDA)~\cite{Blei2003} for texts, 
and Gaussian mixture model (GMM) for  
numeric data. 
Let us consider network change detection problem, 
for example. 
Conventionally, 
this problem has been considered 
on the basis of the (quasi-) difference 
between probability distributions
(e.g., \cite{Hinkley1970,Basseville1993}). 
In other words, 
if the distance between the distributions 
before any given point and that after the point is significantly large, 
then 
we consider the point as a change point. 
Spectral information such as eigenvectors of the association matrix may be employed 
instead
of the distribution difference 
(e.g., \cite{Ide2004,Hirose2009,Akoglu2010}). 
However, 
it is not clearly understood where the change comes from 
with such methods. 
Actually, there are several levels in network changes. 
Let us consider a time-evolving SBM~\cite{Snijders1997,Funke2019} as an example. 
1) On the first level of changes, the connection between the nodes changes. 
2) On the second level, a block distribution changes. 
3) On the third level, the number of blocks or community organization changes. 
We say that the level is higher in the order of 1) $<$ 2) $<$ 3). 
Even if we detect change points with distance-based or spectrum-based methods, 
we can not identify which level the change comes from 
due to the complex nature of networks. 

However, 
it is important to know the levels of the changes. 
The main reason is that we can {\em interpret the cause of a change} 
by looking at its level. 
For example, 
1) corresponds to a change in degrees of communication between nodes. 
2) corresponds to a change in community distribution.  
3) corresponds to a drastic change in community organization. 
   The level of severity of change is in the order of 3) $>$ 2), 1).  
We call this problem of identifying the causes of changes 
the {\em interpretability problem}. 
Note that when 3) occurs, 
changes in degrees of communication between nodes 
and community distribution occur consequently. 
We distinguish this case from 1) and 2), where community organization 
does not change. 

Another reason is that when changes at a higher level occur gradually, 
{\em we can detect signs of the changes by detecting changes at lower levels}. 
For example, the detection of changes at levels 1) and 2) may lead to early signals of changes at level 3). 
We call this the {\em change sign detection problem}.

\begin{figure}[h]
\begin{center}
\includegraphics[width=\linewidth]{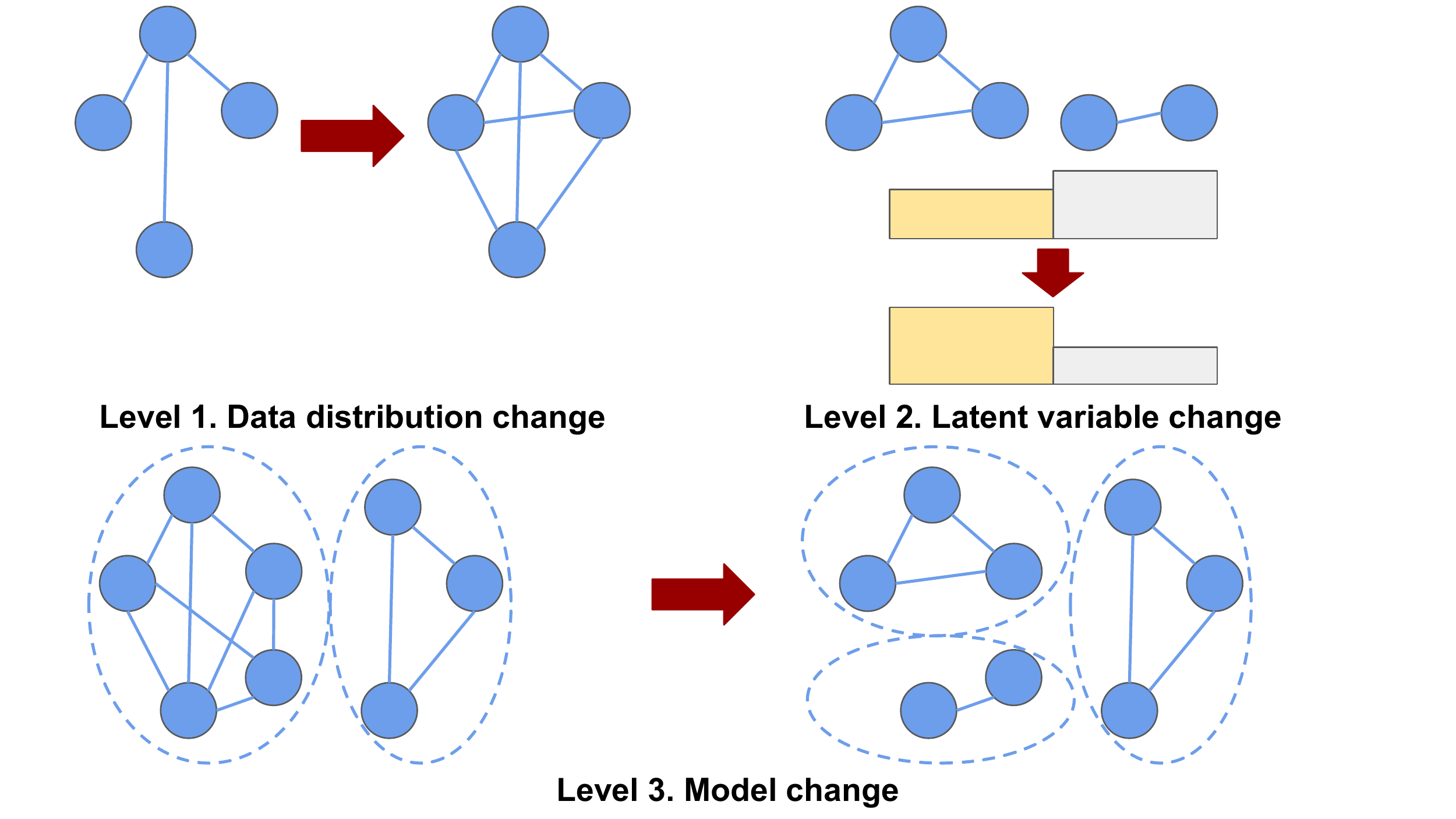}
\caption{Hierarchical changes in latent variable models.}
\label{HCDLfig}
\end{center}
\end{figure}

The above argument is valid for general classes of latent variable models as shown in Fig.~\ref{HCDLfig}. 
1) corresponds to the data distribution change for a fixed latent variable, 
2) corresponds to the change of latent variables for a fixed number of blocks, 
and 3) corresponds to the change of the number of latent variables. 

The primary 
purpose of this paper is to propose a framework 
in which we do not only detect changes in latent variable models 
but also identify their levels in a hierarchical way. 
We name this framework 
{\em hierarchical change detection for latent variable models} (HCDL). 
We present a theory for making 
reliable alarms of changes in a hierarchical way. 
We also 
focus on 
SBM to experimentally demonstrate 
the effectiveness of 
HCDL, 
in terms of change level identification as well as change detection. 

\subsection{Novelty and Significance} 

The novelty and significance of this study are summarized as follows: 

1)	{\em A novel framework for hierarchical change detection. } 
Our HCDL framework is novel in its ability to detect changes at different levels: 
distribution, latent variable, and model. 
For SBM, 
these correspond to individual communication level, 
community distribution one, 
and community organization one, respectively. 
This enables us to identify the cause of any change 
and to interpret the meaning of the change.

Our framework is designed from an information-theoretic viewpoint 
based on the minimum description length~(MDL) principle~\cite{Rissanen1978,
Rissanen2012}. 
That is, 
learning and change detection are conducted 
by finding the probability distributions with the shortest description lengths.
The key idea of our framework 
is to combine the MDL change statistics\cite{Yamanishi2016,Yamanishi2018}
with the DNML (decomposed normalized maximum likelihood) code-length\cite{Wu2017,Yamanishi2019} calculation method 
in change scoring for latent variable models.  
First, 
we employ the  MDL change statistics 
to score the degree of change at any given time point. 
It measures how much the data can be compressed by changing the model at that time point. 
In the original MDL change statistics, 
NML code-length~\cite{Rissanen2012} has been employed 
to calculate code-lengths. 
When a latent variable model is considered, 
however, 
hierarchical changes can not be detected 
if we simply apply the NML code-length.  
This is because the NML code-length can not be decomposed 
for data and latent variables. 
Thus, 
we propose to use the DNML code-length 
instead of the NML code-length. 
Therefore, we can identify from which level any change comes.

2)	{\em Theoretical basis for making reliable alarms.} 
The test for changes based on the MDL change statistics with DNML code-length 
is referred to as the \textit{MDL test} throughout this study. 
In order to make reliable alarms for the MDL test, 
we need to determine the threshold for change scores at respective levels. 
We theoretically derive upper bounds on Type I and Type 
II error probabilities for the MDL test and thereby design the thresholds for scores so that Type I error probability (false alarm rate) 
is properly upper bounded. 
The thresholds are determined for all levels (distribution, 
latent variable, and model). 
Hence, 
we 
could raise alarms of hierarchical changes 
for which the reliability is theoretically guaranteed.

3)	{\em Novel analysis of hierarchical changes in terms of change interpretation and change sign detection.}
Through 
HCDL, 
we offer two new methodologies for analyzing change detection. 
One is {\em hierarchical interpretation of detected changes.} With HCDL, 
we can not only detect the changes 
but also identify from which level 
the changes come from. 
Changes may simultaneously 
originate from different levels. 
Then, 
we can quantitatively 
analyze 
how much the individual changes 
contribute to the overall one. 
This yields a new approach to interpretation of changes. The other methodology is {\em change sign detection. } 
Suppose that a change occurs gradually. 
Then, 
we expect that changes at lower levels may occur before those at higher ones. 
For example, 
in the case of network change detection, 
changes in network connections 
among given communities or 
those of the community distribution 
may be 
signs of a more drastic community structure change. 
We conduct experiments with 
synthetic and real datasets to demonstrate the effectiveness 
of change detection at respective levels, 
hierarchical interpretability, 
and change sign detection with HCDL.

\subsection{Related Work}

The issue of change detection 
in latent variable models 
such as network models 
has been explored extensively 
in the areas of knowledge discovery and data mining 
(
 e.g., 
\cite{Kim2018,Akoglu2015,Ranshous2015}
). 
Most of previous studies were concerned with detection of 
parameter changes in data distributions. 
Such studies often employed distribution-difference-based 
or spectrum-based methods, as in Section~\ref{subsection:motivation_hcdl}. 
However, these methods can not deal with changes at higher levels such as structural changes. 
Hence, the different levels of changes can not be identified. 
Therefore, these approaches lack interpretability of changes. 
Some previous studies were concerned with 
the issue of dynamic model selection~\cite{Yamanishi2007}. 
See also related notions such as 
tracking best experts~\cite{Herbster1998} 
and switching distributions~\cite{Erven2012}. 
Some previous studies were concerned with 
the issue of dynamic model selection\cite{Yamanishi2007}. 
Time-evolving networks~\cite{Song2005} and 
Graphscope~\cite{sun2007} dealt with 
topics related to dynamic model selection 
(see also the survey \cite{Rossetti2018} in community detection in dynamic networks). 
That is, 
Their goal is to detect changes 
in discrete structures,  
the number of clusters, 
and 
the number of communities 
at higher levels. 
However, 
among such studies, 
there was no attempt to identify changes at lower levels as well as those of higher ones. 

The main target of conventional 
studies on change detection 
was abrupt change. 
Recently, gradual change 
\cite{Yamanishi2016} has become a new target of change detection. 
Incremental changes have also been discussed in the scenario of concept drifts~\cite{Gama2014}. 
It is a challenging new problem: 
how can we detect signs of changes when they are gradual or incremental? 
There are a number of 
studies on model change sign detection~\cite{Huang2014,
Hirai2018,Hirai2019,Ohsawa2018,Fukushima2019}. 
However, 
the change signs studied there 
have not been related to changes at different levels; therefore, the causes of changes in signs can not be explained.

%% file: 2_Problem_Setting.tex
\section{Problem Setting}
\label{section:problem_setting}

In this section, 
we describe the problem setting of change detection 
in latent variable models.  
Let ${\mathcal X}$ be the data domain, 
which can be either continuous or discrete, 
and ${\mathcal Z}$ be the range of latent variables, 
which is finite.
Let $X\in {\mathcal X}$ and $Z\in {\mathcal Z}$ be random variables, 
and $x$ and $z$ denote their realizations, respectively. 
We denote $x_{a}^{b} = x_{a} \dots x_{b}$. 
When $a=1$, we write $x_{1}^{b}$ as $x^{b}$.
We write $x_{1}^{b}$ as $x^{b}$.

Let ${\mathcal M}$ be the set of discrete models, e.g., clusters and communities.
Suppose that 
each datum $(x_{t}, z_{t})$ at time $t$ 
is independently drawn from 
a probability density (or mass) function 
of a latent variable model   
of the following general form: 
\begin{align}
(x_{t}, z_{t}) \sim f(X, Z; \theta, M), 
\label{eq:probability_density_function_latent_variable_model}
\end{align}
where $M \in \mathcal{M}$ is a model. 
The probability density (or mass) function 
in Eq.~\eqref{eq:probability_density_function_latent_variable_model} 
is assumed to be factorized as 
\begin{align}
f(X, Z; \theta, M) = f(X | Z; \theta_{1}, M) f (Z; \theta_{2}, M),
\label{eq:probability_density_function_latent_variable_model_decomposed} 
\end{align}
where $\theta =(\theta _{1}, \theta _{2})$, $\theta_{1}$ is the real-valued parameter of 
probability density function of 
observed variables given latent variables, while
$\theta_{2}$ is the real-valued parameter of 
the probability mass function of  latent variables. 
In the case of 
SBM, 
$X$ is a data matrix, $Z$ is the block assignment of each data, $\theta_{1}$ is an edge weight parameter, $\theta_{2}$ is a parameter of block distribution, and $M$ is a structure indicating 
the number of blocks. 

We are concerned with 
detecting the three levels in changes: 
\setlist{nolistsep}
\begin{itemize}
\itemsep0em 
\item {\em Level 1: Change of parameter $\theta _{1}$ of data distribution for fixed latent variables}.
It means an inner-level change in the probabilistic model for $X$ for given $Z$. For example, changes in connections within a community in SBM belong to level 1.
\item {\em Level 2: Change of parameter $\theta_{2}$ of latent variable distribution}. 
It means a change in the probabilistic model for $Z$. 
For example, 
community distribution  change in SBM belongs to level 2.
\item {\em Level 3: Change of model M}. 
It means a drastic structural change in the probabilistic model for $X$ and $Z$. 
For example, change in the number of communities or community organization in SBM belongs to level 3.
\end{itemize}

Our target issues are summarized as follows:
\setlist{nolistsep}
\begin{enumerate}
\itemsep0em 
\item How can we detect changes at each level accurately?
\item How can we interpret the detected changes by relating them to their levels?
\item How can we find signs of changes at higher levels by detecting changes at lower levels?
\end{enumerate}

We address these issues using the HCDL framework.

%% file: 3_Information_Theoretic_Methods_for_Change_Detection.tex
\section{Information Theoretic Methods for Change Detection}
\label{information_theoretic_methods_for_change_detection}

We take an information-theoretic approach 
to change detection on the basis of the MDL principle.
That is, 
we detect a change point by finding a time point 
so that the total code-length required for encoding the data and its distribution is shortest.
Two key notions 
to realize the hierarchical change detection 
for latent variable models 
are the MDL change statistics 
and the DNML code-length. 
We effectively combine these two notions in our HCDL framework.
In this section, 
we focus on the 
issue of detecting changes at level 3 
in order to 
clarify the information-theoretic methodology.

\subsection{MDL Change Statistics}
\label{subsection:sequential_dnml_change_detection}

The MDL change statistics~\cite{Yamanishi2018,Yamanishi2016} 
is a score that measures the degree of change 
for a specified time point $t$ within a given window. 
Let $t$ be a given time point. 
Prepare a window of size $2h$ whose center is $t$ ($h \in \mathbb{N})$.
The MDL change statistics at time $t$, which we denote as $\Phi_{t}$,  
is defined as the difference between 
the total code-length required for encoding the data 
with an unchanged model and that with models changed 
before and after time $t$.
It is formally defined as follows: 

\begin{definition}{\textrm (MDL change statistics)}
For a fixed window size $2h \in \mathbb{N}$,
let ${\bm x}_{(t)}=x_{t-h+1}^{t+h}$,
${\bm x}_{(t)}^{-}=x_{t-h+1}^{t}$, ${\bm x}_{(t)}^{+}=x_{t+1}^{t+h}$.
Similarly, ${\bm z}_{(t)},{\bm z}^{+}_{(t)},{\bm z}^{-}_{(t)}$ are defined.
The \textit{MDL change statistics} $\Phi_{t}$ is defined as
\begin{align}
\Phi_{t}
&= \frac{1}{2h}
   \min_{M}
     \left\{
       L({\bm x}_{(t)}, {\bm z}_{(t)}; M)
       + L (M)
     \right\}   \\
&\quad
   -\frac{1}{2h}
    \min_{M_{1}, M_{2}}
      \biggl\{
        L({\bm x}_{(t)}^{-}, {\bm z}_{(t)}^{-}; M_{1}) +
        L({\bm x}_{(t)}^{+}, {\bm z}_{(t)}^{+}; M_{2}) \nonumber \\
&\quad \quad \quad \quad \quad \quad \quad 
        +
        L (M_{1}, M_{2})
      \biggr\}. 
\label{eq:def_SDNML}
\end{align}
Here, 
$L({\bm x}, {\bm z}; M)$ denotes the code-length 
for $({\bm x},{\bm z})$ relative to model $M$. 
$L(M)$ is the code-length for model $M$, 
whereas $L(M_{1}, M_{2})$ is the code-length 
required for encoding model $M_{1}$ and $M_{2}$.
\end{definition}
The code-length $L(w)$ for $w$ means 
the length of codewords when $w$ is encoded into 
a binary sequence under the prefix condition 
that any codeword is not a prefix of any other ones.
This condition holds if and only if Kraft's inequality holds: 
$\sum _{w}2^{-L(w)}\leq 1$, 
where the sum ranges over all possible $w$s.
Intuitively, 
the MDL change statistics measures the degree of change 
at time $t$ in terms of how much data is compressed 
by using different models before and after time $t$.
We describe how to calculate $L(M)$ and $L(M_{1}, M_{2})$ in Section~\ref{subsection:synthetic_dataset_1}.

When a data stream is given, 
we slide the window to sequentially calculate
Eq.~\eqref{eq:def_SDNML}. 
We thereby obtain a sequence of the MDL change statistics. 
By finding a time point when the MDL change statistics 
exceeds a threshold, 
we can find a change point of model $M$, i.e., at level 3.

In the original definition of the MDL change statistics, 
the \textit{normalized maximum likelihood}~(NML) code-length is employed 
to calculate $L({\bm x},{\bm z};M)$ in it.
It is formalized as
\begin{align}
L_{_\mathrm{NML}}({\bm x},{\bm z}; M)
&=-\log f({\bm x},{\bm z}; \hat{\theta}({\bm x},{\bm z}), M)  \nonumber \\
&\quad 
  +\log \sum _{{\bm x}',{\bm z}'} f({\bm x}',{\bm z}'; \hat{\theta}({\bm x}',{\bm z}'), M),
\label{eq:NML}
\end{align}
where $\hat{\theta}({\bm x},{\bm z})$ is the maximum likelihood estimator of $\theta $ from $({\bm x},{\bm z})$.

The problem in the NML code-length is that it can not be decomposed into the part of data and that of latent variables.
It can not be decomposed into 
the part of data and that of latent variables. 
Hence, we can not identify the three levels of changes as in 
Section~\ref{section:problem_setting} if we simply employ the NML code-length in the MDL change statistics.

\subsection{DNML Code-Length}

In order to resolve the problem in the previous section, we employ 
the \textit{decomposed normalized maximum likelihood} (DNML) code-length for latent variable models 
instead of the NML code-length.
In calculating the DNML code-length, 
the total code-length is decomposed into the sum of the NML code-length for data conditioned on latent variables and that for the latent variables.  
According to \cite{Yamanishi2019,Wu2017}, 
it is formulated as follows:
\begin{definition}{(DNML code-length)} \cite{Yamanishi2019,Wu2017} 
The DNML code-length for $(x, z)$ for a model $M$ is defined as 
\begin{align}
L_{_\mathrm{DNML}}(x, z; M) \mydef  
L_{_\mathrm{NML}}(x | z; M) + L_{_{\mathrm{NML}}}(z; M), 
\label{eq:def_dnml}
\end{align}
where 
\begin{align}
L_{_\mathrm{NML}}(x | z; M)
&= -\log{ f(x | z; \hat{\theta}_{1}(x, z); M) }
   +\log{ C_{X|z}(M) }, \\
L_{_\mathrm{NML}}(z; M)
&= -\log{ f(z; \hat{\theta}_{2}(z), M ) }  
   +\log{ C_{Z}(M) }. \nonumber
\end{align}
Here, 
$\hat{\theta}_{1}$ and $\hat{\theta}_{2}$ 
are the maximum likelihood estimators of 
$\theta_{1}$ and $\theta_{2}$ as in Eq.~\eqref{eq:probability_density_function_latent_variable_model_decomposed}, 
and 
\begin{align}
C_{X |z}(M)
&\mydef \sum_{x} f( x | z; \hat{\theta}_{1}(x, z), M), \\ 
\setlength{\belowdisplayskip}{1pt} 
C_{Z}(M)
&\mydef \sum_{z} f(z; \hat{\theta}_{2}(z), M). 
\label{eq:parametericcomplexity}
\end{align}
\end{definition}

According to \cite{Yamanishi2019,Wu2017}, 
it is known that 
$C_{X|z}(M)$ and $C_{Z}(M)$ are efficiently computable in order $O(n+|{\mathcal M}|)$ 
for a wide range of classes of latent variable models 
such as 
SBM, LDA, and GMM. 

\begin{example}{\textrm (DNML for SBM)}
SBM is a canonical model
for community detection \cite{Snijders1997}.
SBM partitions the vertices of a network
into groups.
We assume that every group has
its own probability to generate a link.
A model $M$ denotes a structure of partitioning.

According to \cite{Wu2017}, the DNML code-length $L_{_\mathrm{DNML}}(x, z; M)$ for SBM is calculated as follows:
\begin{align} 
L_{_\mathrm{DNML}}(x, z; M)
&= \sum_{k_{1}} \sum_{k_{2}}
     \left(
       n_{k_{1} k_{2}}
       \log{ n_{k_{1} k_{2}} }
     \right.  \nonumber \\
&\quad \quad \quad 
     \left.
       -
       n_{k_{1} k_{2}}^{+}
       \log{ n_{k_{1} k_{2}}^{+} }
       -
       n_{k_{1} k_{2}}^{-}
       \log{ n_{k_{1} k_{2}}^{-} }
     \right)  \nonumber \\
&\quad
   +\sum_{k_{1} } \sum_{k_{2}}
      \log{ C(n_{k_{1} k_{2} }, 2) }  \nonumber \\
&\quad 
   +\sum_{k} n_{k} (\log{n} - \log{n_{k}}) 
   +\log{ C(n, K) }, 
\label{eq:dnml_sbm}
\end{align}
where
$n_{k_{1} k_{2}}^{+}$ and
$n_{k_{1} k_{2}}^{-}$ are
the number of links and no-links in group $(k_{1}, k_{2})$.
$n_{k_{1} k_{2}}$ is defined as
$n_{k_{1} k_{2}} =n_{k_{1} k_{2}}^{+} + n_{k_{1} k_{2}}^{-}$,
which is the total number of
links in group $(k_{1}, k_{2})$.
$\log{ C (n, K)}$ is the normalization term $C_{Z}$ as in Equation~\eqref{eq:parametericcomplexity} for data of length $n$ following
the multinomial distribution with $K$ elements.
\end{example}

Since $z$ is not observable, 
we may instead 
employ the estimator of $\hat{z}(x)$ from $x$ 
instead of $z$. 
As an estimator, for example, 
we may employ the one that maximizes 
the posterior probability estimated by the Expectation-Maximization (EM) algorithm. 

In the following, 
we denote 
$\bm{x}_{(t)}=x_{t-h+1}^{t+h}$, 
$\bm{x}^{-}_{(t)} = x_{t-h+1}^{t}$, 
$\bm{x}^{+}_{(t)} = x_{t+1}^{t+h}$.  
Similarly, 
$\bm{z}_{(t)}$, 
$\bm{z}^{-}_{(t)}$, 
and 
$\bm{z}^{+}_{(t)}$ 
are defined likewise. 
The DNML can be applied to model selection~\cite{Yamanishi2019}. 
That is, the DNML model estimator $\hat{M}_{t}$ from $({\bm x}_{(t)},{\bm z}_{(t)})$ at time $t$ is given by
\begin{align} 
\hat{M}_{t}=\argmin_{M} \{L_{_\mathrm{DNML}}({\bm x}_{(t)},\hat{{\bm z}}_{(t)};M)+L(M)\}, 
\label{eq:DMDLestimator}
\end{align}
where $L(M)$ denotes the code-length for model $M$, 
and 
$L_{_\mathrm{DNML}}(\bm{x}_{(t)}, \bm{z}_{(t)}; M)$ is calculated 
as the sum of DNML code lengths in the window:  
\begin{align}
L_{_\mathrm{DNML}}(\bm{x}_{(t)}, \bm{z}_{(t)}; M)
&= \sum_{\tau=t-h+1}^{t+h}
     L_{_\mathrm{DNML}}(x_{\tau}, z_{\tau}; M), 
\end{align}
where $\hat{\theta}_{1}$ and $\hat{\theta}_{2}$ are estimated 
to minimize the log-likelihoods: 
\begin{align} 
\hat{\theta}_{1}(\bm{x}_{(t)}, \bm{z}_{(t)})
&= \argmin_{\theta_{1}} 
     \sum_{\tau=t-h+1}^{t+h} 
       -\log{ 
            f (x_{\tau} | z_{\tau}; \theta_{1}, M) 
       }, 
\label{eq:estimated_theta1_in_window} \\
\hat{\theta}_{2}(\bm{z}_{(t)})
&= \argmin_{\theta_{2}}
     \sum_{\tau=t-h+1}^{t+h}
       -\log{ 
            f (z_{\tau}; \theta_{2}, M) 
       }.
\label{eq:estimated_theta2_in_window}
\end{align}
Plugging the DNML formula in Eq.~\eqref{eq:def_dnml} 
into the formula of the MDL change statistics 
yields  
the MDL change statistics combined with DNML as follows:
\begin{align}    \label{eq:def_MDL+DNML}
\Phi_{t}
&= \frac{1}{2h}
   \min_{M}
     \left\{
       L_{_\mathrm{DNML}}({\bm x}_{(t)}, {\bm z}_{(t)}; M)
       + L (M)
     \right\}  \nonumber \\
&\quad
   -\frac{1}{2h}
    \min_{M_{1}, M_{2}}
      \left\{
        L_{_\mathrm{DNML}}({\bm x}_{(t)}^{-}, {\bm z}_{(t)}^{-}; M_{1})  
      \right.  \nonumber \\
&\quad \quad 
      \left.
        +
        L_{_\mathrm{DNML}}({\bm x}_{(t)}^{+}, {\bm z}_{(t)}^{+}; M_{2})
        +
        L (M_{1}, M_{2})
      \right\}. 
\end{align}

By sliding the window, 
for a threshold parameter 
$\epsilon >0$, 
we determine that a model change 
occurred 
if $\Phi _{t}>\epsilon$ 
and that it 
did not, 
otherwise. 
We call this test the \textit{MDL test}.
Once the model change is detected, we can identify the model 
with 
Eq.~\eqref{eq:DMDLestimator}.

%% file: 4_Theory_for_Making_Reliable_Alarms.tex
\section{Theory for Making Reliable Alarms}
\label{section:theory_for_making_reliable_alarms}

\subsection{Error Probabilities for Change Detection}

We need to choose a threshold parameter 
$\epsilon$ 
for the MDL change statistics 
in order to make reliable alarms. 
to make reliable alarms. 
For this purpose, 
we first present a theoretical property of 
the MDL change statistics 
in terms of error probabilities 
in the scenario of hypothesis testing.
We then derive the threshold 
so that the error probabilities are properly bounded.
We continue to focus on the change detection of level 3, 
i.e., 
model change detection.

Let us consider 
the following hypothesis testing.
For a 
time point $t$, 
the null hypothesis $H_{0}$ is that 
no model change occurs at time $t$, 
while the alternative hypothesis $H_{1}$ is that it occurs 
at time $t$. 
Here, we do not know anything about model $M$.
\begin{align*}
H_{0} &: 
  ({\bm x}_{(t)}, {\bm z}_{(t)}) 
  \sim f(X^{2h}, Z^{2h};     
         \theta^{\ast}_{0},
         M^{\ast}_{0}), \\
H_{1} &: 
  ({\bm x}^{-}_{(t)}, {\bm z}_{(t)}^{-}) 
  \sim f(X^{h}, Z^{h};
         \theta^{\ast}_{1},
         M^{\ast}_{1}),  \\
  &\quad  
  ({\bm x}^{+}_{(t)},{\bm z}_{(t)}^{+}) 
  \sim f(X^{h}, Z^{h};
         \theta^{\ast}_{2}, 
         M^{\ast}_{2}), 
\end{align*}
where 
$M^{\ast}_{0}$, $M^{\ast}_{1}$, and $M^{\ast}_{2}$ ($M^{\ast}_{1}\neq M^{\ast}_{2}$)
are the unknown true models, 
and $\theta^{\ast}_{0}$, $\theta^{\ast}_{1}$, and $\theta^{\ast}_{2}$ 
are the unknown true values of the parameters. 
 
We conduct the hypothesis testing using 
the MDL test as follows: 
we accept $H_{1}$ 
if $\Phi _{t}>\epsilon$ for 
$\Phi_{t}$ as in  
Eq.~\eqref{eq:def_MDL+DNML};
otherwise, 
we accept $H_{0}$. 
We evaluate this test 
in terms of Type I and II error probabilities. 
For the Type I error probability, 
$H_{0}$ is true, 
but $H_{1}$ is accepted by the MDL test. 
Meanwhile, 
for the Type II error probability, 
$H_{1}$ is true, 
but $H_{0}$ is accepted by the MDL test.
We have the following theorem on the MDL test for model changes. 

\begin{theorem}\label{theorem:error_probabilities_for_dnml_change_test}
The Type I error probability $\delta_{1}$ 
and Type II error probability $\delta_{2}$ 
for the MDL test 
with $\Phi_{t}$ in Eq.~\eqref{eq:def_MDL+DNML}
are given as follows:
\begin{align}
&\delta_{1} 
\leq \exp{
   \left\{
     -2h \left(
       \epsilon - 
       \frac{ \log{ C(M_{0}^{\ast})  } 
              +  L (M_{0}^{\ast})  }{2h}
     \right)
   \right\}
  },  \label{eq:type1}\\
&\delta_{2} \leq
  \exp{
    \left\{
      -n \left(
        d( {f}_{ _{\mathrm{NML}} }, 
                   f_{M_{1 \ast 2}} 
        ) 
        -
        \frac{
          \ell (M_{1}^{\ast}, M_{2}^{\ast}, \epsilon)
        }{
          2n
        }
      \right)
    \right\}
  }, \label{eq:type2}
\end{align}
where 
\begin{align}
&\log{C(M)} 
\mydef \log{ C_{X|Z}(M)C_{Z}(M) }, 
\, 
C_{X|Z}(M) 
\mydef \max _{{\bm z}}C_{X|{\bm z}}, \nonumber \\
&\ell (M_{1}^{\ast}, M_{2}^{\ast}, \epsilon)
\mydef 
        \log{ C( M_{1}^{\ast} ) 
              C( M_{2}^{\ast} )  }
 + L ( M_{1}^{\ast}, M_{2}^{\ast} )
 + 2h\epsilon,  \nonumber \\
&d (f_{_\mathrm{NML}}, f_{M_{1 \ast 2}})  \nonumber \\
&\mydef -\frac{1}{n} 
 \log
   \sum_{{\bm x}_{(t)}, 
         {\bm z}_{(t)}}  
   \left(
     f_{_\mathrm{NML}}({\bm x}_{(t)},{\bm z}_{(t)}) 
     f({\bm x}_{(t)},{\bm z}_{(t)}; M_{1} \ast M_{2})
   \right)^{\frac{1}{2}}, \nonumber \\
&f_{_\mathrm{NML}}({\bm x}_{(t)},{\bm z}_{(t)}) 
\mydef 
 \frac{
   2^{-\min _{M}L_{_\mathrm{NML}}({\bm x}_{(t)},{\bm z}_{(t)};M)} 
 }{
   \sum _{{\bm x},{\bm z}}2^{-\min _{M}L_{_\mathrm{NML}}({\bm x},{\bm z}; M)}
 }, \nonumber \\
&f(\bm{x}_{(t)}, \bm{z}_{(t)}; 
  M_{1} \ast M_{2})  \nonumber \\
&\mydef f(\bm{x}_{(t)}^{-},
   \bm{z}_{(t)}^{-}; \theta^{\ast}_{1}, M_{1}^{\ast})  
 f(\bm{x}_{(t)}^{+},
   \bm{z}_{(t)}^{+}; \theta^{\ast}_{2}, M_{2}^{\ast}). \nonumber
\end{align}
\end{theorem}
Theorem~\ref{theorem:error_probabilities_for_dnml_change_test} can be proven by extending Theorem 3.1 in \cite{Yamanishi2018} 
to 
latent variable models. 
Theorem~\ref{theorem:error_probabilities_for_dnml_change_test} shows that both Type I and II error probabilities 
for the MDL test with the DNML code-length 
converge to zero exponentially as the sample size increases if $\epsilon$ is properly designed.
The rates of convergence depend on the information complexity of true models and the discrepancy measure of the probability distributions before and after the change.

\subsection{Choosing Threshold Parameters}

On the basis of the theory in the previous section, 
we show how to 
determine the threshold parameter $\epsilon$ 
to make a reliable alarm of a change. 
The key idea is to determine $\epsilon$ so that Type I error probability is upper-bounded by a predetermined confidence parameter.
That is, 
letting a confidence parameter be $\delta > 0$, 
we choose $\epsilon $ so that the Type I error probability in Eq.~\eqref{eq:type1} is upper-bounded by a confidence parameter $\delta$. 
This implies that 
$\epsilon $ should satisfy the following inequality: 
\begin{align}
\epsilon
\geq 
  \frac{
    \log{ C(M_{0}^{\ast}) } 
    + \log{ L(M_{0}^{\ast}) } - \log{\delta} 
  }{
    2h
  }. 
\label{eq:threshold_inequality_for_model}
\end{align}
We use this criterion to choose $\epsilon$ 
for given $\delta$ 
in order to guarantee the reliability of 
an alarm. 
Since $\delta$ is a much smaller value than $\epsilon$, 
the choice of $\delta$ does not affect the result compared to that of $\epsilon$.

Note that the true model $M_{0}^{\ast}$ in Eq.~\eqref{eq:threshold_inequality_for_model} 
is not known in real cases.
Hence, we estimate it as $\hat{M}_{t}$ 
in 
Eq.~\eqref{eq:DMDLestimator} and plug it into 
Eq.~\eqref{eq:threshold_inequality_for_model}. We adopt the right-hand side value of Eq.~\eqref{eq:threshold_inequality_for_model} as $\epsilon$.

%% file: 5_Hierarchical_Change_Detection.tex
\section{Hierarchical Change Detection} 
\label{section:hierarchical_change_specification_algorithm}

In Sections~\ref{information_theoretic_methods_for_change_detection} 
and \ref{section:theory_for_making_reliable_alarms}, 
we focused on detecting changes at level 3.
This section extends the methodology into the hierarchical change detection for all three levels. 

Let us denote $\hat{M}$ as the minimizer 
of the first term of the right-hand side of the MDL change statistics
in Eq.~\eqref{eq:def_MDL+DNML}, 
and $\hat{M}_{1}, \hat{M}_{2}$ 
as the minimizers in the second term. 
In order to extend our discussion in previous sections into hierarchical change detection, 
we first 
note that $\Phi _{t}$ in Eq.~\eqref{eq:def_MDL+DNML} is decomposed as follows:
\begin{align}
\Phi_{t}
&= \Phi_{t}^{X|Z}
   + \Phi_{t}^{Z}
   + \Delta L_{t} ( \hat{M}, \hat{M}_{1}, \hat{M}_{2} ), 
\label{eq:decomposition}
\end{align}
where
\begin{align*}
\Phi_{t}^{X|Z}
&\mydef 
 \frac{1}{2h} 
   \left\{
     L_{_\mathrm{NML}} ({\bm x}_{(t)} | {\bm z}_{(t)}; \hat{M})
   \right. \\
&\quad \quad 
   \left.
     -(L_{_\mathrm{NML}} ({\bm x}_{(t)}^{-}| {\bm z}_{(t)}^{-}; \hat{M}_{1})
     +L_{_\mathrm{NML}}({\bm x}_{(t)}^{+} | {\bm z}_{(t)}^{+}; \hat{M}_{2}) ) 
   \right\},  \\
\Phi_{t}^{Z}
&\mydef
\frac{1}{2h} 
  \left\{
    L_{_\mathrm{NML}} ({\bm z}_{(t)}; \hat{M}) 
  \right.  \\
&\quad \quad 
  \left. 
    -(L_{_\mathrm{NML}} ({\bm z}_{(t)}^{-}; \hat{M}_{1})
    +L_{_\mathrm{NML}} ({\bm z}_{(t)}^{+}; \hat{M}_{2}))
  \right\}, \\
\Delta L_{t} 
&\mydef 
 \frac{1}{2h} \left\{
   L(\hat{M}) - L(\hat{M}_{1}, \hat{M}_{2})
 \right\}. 
\end{align*}
Eq.~\eqref{eq:decomposition} means that the MDL change statistics can be decomposed into the sum of those for different levels. 
$\Phi_{t}^{X|Z}$ is the MDL change statistics for data $X$ given latent variable $Z$, while $\Phi_{t}^{Z}$ is that for $Z$ itself.
This decomposition owes to the nature of DNML code-length.

We can realize hierarchical change detection 
by making use of the property in 
Eq.~\eqref{eq:decomposition} as follows:
Let $\epsilon$ be a threshold parameter satisfying Eq.~\eqref{eq:threshold_inequality_for_model}.  
If $\Phi_{t}>\epsilon$, then we determine that a model change 
occurred, 
i.e., $\hat{M} \neq \hat{M}_{1}$ or 
$\hat{M} \neq \hat{M}_{2}$.
Otherwise, we determine that no model change 
occurred, i.e.,
$\hat{M} = \hat{M}_{1} = \hat{M}_{2}$.

Then, we further investigate whether any change 
occurred at lower levels.
Let $\epsilon _{_{X|Z}} > 0$ and $\epsilon _{_{Z}} > 0$ be 
threshold parameters. 
If $\Phi^{Z}_{t} > \epsilon_{_{Z}}$ holds, 
we determine that a change 
occurred  
in the distribution over 
latent variable $Z$. 
Otherwise, we determine that 
such a change did not occur (level 2).
Likewise, 
if $\Phi^{X|Z}_{t} > \epsilon_{_{X|Z}}$, 
we determine that a change 
occurred with respect to 
observed variable $X$ 
for given latent variable $Z$.
Otherwise, 
we determine that such a change did not occur (level 1).

The threshold parameters $\epsilon_{_{X|Z}}$ and $\epsilon_{_{Z}}$ are determined similarly. 
That is, 
letting $\delta_{_{X|Z}}>0$ and $\delta_{_{Z}}>0$ be given confidence parameters, 
$\epsilon_{_{X|Z}}$ and $\epsilon_{_{Z}}$ are determined so that Type I error probability for the MDL test is upper-bounded by $\delta_{_{X|Z}}$ and $\delta_{_{Z}}$, respectively. 
That is, for model $M$, we have
\begin{align}
\epsilon _{_{X|Z}}
\geq \frac{ \log{C_{X|Z}(M)} - \log{\delta_{_{X|Z}}} }{2h}, 
\epsilon_{_{Z}}
\geq \frac{ \log{C_{Z}(M)} - \log{\delta_{_{Z}}} }{2h},
\label{eq:estimation_of_threshold_for_changes_of_parameters}
\end{align}
where $C_{X|Z}(M)$ and $C_{Z}(M)$ are as in Eq.~\eqref{eq:parametericcomplexity}. 

Note that the severity of levels 1 and 2 can not be ordered linearly 
when changes at level 1 and 2 occur simultaneously, 
we can evaluate the importance of both, 
respectively, 
as follows:
\begin{align}
w_{t}^{X|Z}
&= \frac{ \Phi_{t}^{X|Z} }{ \Phi_{t}^{X|Z} + \Phi_{t}^{Z} }, 
\quad 
w_{t}^{Z}
= \frac{\Phi_{t}^{Z} }{  \Phi_{t}^{X|Z} + \Phi_{t}^{Z} }.
\label{eq:weights_of_observed_variables_and_latent_variables}
\end{align}

Summarizing the above arguments, 
we show the core algorithm for HCDL in 
Algorithm~\ref{algorithm:hcsd}. 
Through this algorithm, 
we can detect changes at different levels, thereby can interpret the causes of the changes by identifying their levels. 

\setlength{\textfloatsep}{0pt}

\begin{algorithm}[h]
\setstretch{0.95}
\caption{Hierarchical change detection algorithm for latent variable models (HCDL)}
\label{algorithm:hcsd}
\begin{small}
\begin{algorithmic}[1]
\REQUIRE $h$: window size, $\delta, \delta_{_{X|Z}}, \delta_{_{Z}}$: parameters for controlling Type I error probabilities of model and parameter changes.  
\ENSURE 
\FOR {$t=h, h+1,\, \dots$}
  \STATE Calculate the MDL change statistics $\Phi_{t}$. 
  \IF {$\Phi_{t} > \epsilon $}
    \STATE Raise an alarm of change at level 3. 
  \ELSE
    \STATE Estimate the model at $t$ as $\hat{M}_{t}$ according to Eq.~\eqref{eq:DMDLestimator}. 
    \STATE Determine the threshold parameters $\epsilon _{_{X|Z}}$, 
    $\epsilon_{_{Z}}$ as in Eq.~\eqref{eq:estimation_of_threshold_for_changes_of_parameters}. 
    \IF {$\Phi^{X|Z}_{t} > \epsilon_{_{X|Z}}$}
      \STATE Raise an alarm of change at level 1. 
    \ENDIF
    \IF {$\Phi^{Z}_{t} > \epsilon_{_{Z}}$}
      \STATE Raise an alarm of change at level 2. 
    \ENDIF
    \IF {$\Phi^{X|Z}_{t} > \epsilon_{_{X|Z}}, \Phi^{Z}_{t} > \epsilon_{_{Z}}$}
      \STATE Calculate the weights $w_{t}^{X|Z}$ and $w_{t}^{Z}$ 
             of changes at levels 1 and 2 according to
             Eq.~\eqref{eq:weights_of_observed_variables_and_latent_variables}. 
    \ENDIF
  \ENDIF
\ENDFOR
\end{algorithmic}
\end{small}
\end{algorithm}


%% file: 6_Experiments.tex
\section{Experiments}

In this section,
we conduct experiments 
with synthetic and real datasets 
to demonstrate the effectiveness of HCDL
\footnote{
Source codes are available at \url{https://github.com/s-fuku/hcdl}.}. 

\subsection{Synthetic Dataset 1 (Abrupt Change)}
\label{subsection:synthetic_dataset_1}

\subsubsection{Dataset}

\begin{figure}[tb]
\centering
\includegraphics[width=\linewidth]{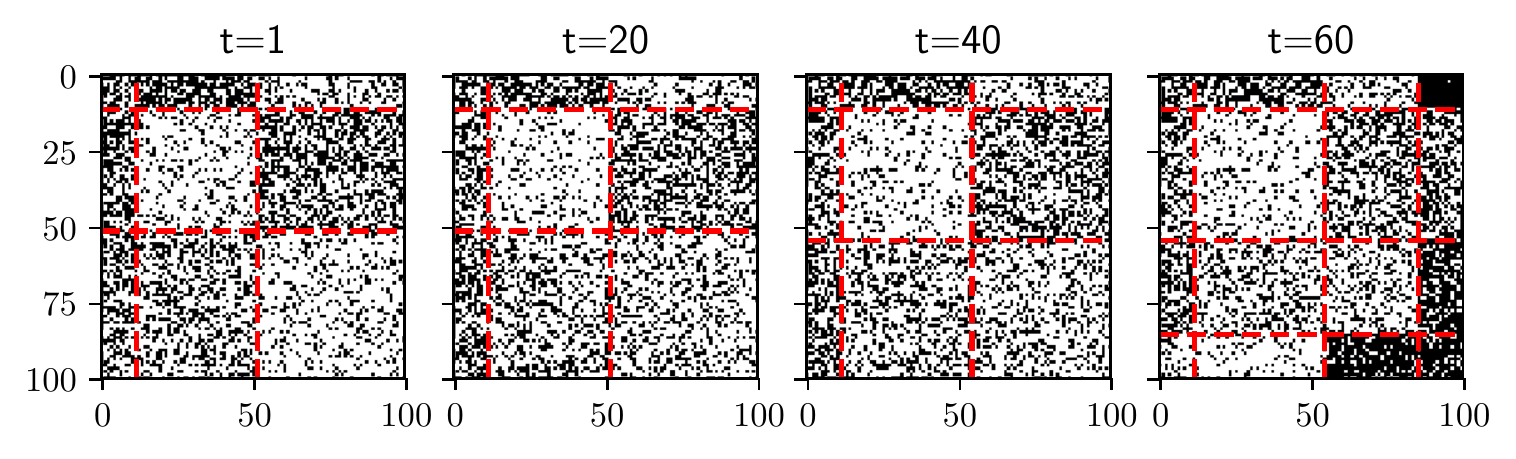}
\caption{ A sample sequence of links $x_{t}$ for abrupt change. 
  At $t=20$, the link probability changes abruptly. 
  At $t=40$, the mixture probability changes abruptly. 
  At $t=60$, the model changes from $K=3$ to $K=4$ abruptly. 
}
\label{fig:sample_X_abrupt}
\end{figure}

For $t=1, \dots, 80$, 
we generated links $x_{t}$ between nodes as follows:
\begin{align}
\renewcommand{\arraystretch}{0.7}
x_{t} \sim 
    \begin{cases}
    \mathrm{SBM} \, (x_{t}; \pi^{1}, \theta^{1}, K=3)
    & (t=1), \\
    \mathrm{LinkTrans} \, (x_{t-1} | z_{t-1}; \theta^{1}, \beta)
    & (t=2, \dots, 19), \\
    \mathrm{SBM} \, (x_{t}; \pi^{1}, \theta^{2}, K=3)
    & (t=20), \\
    \mathrm{LinkTrans} \, (x_{t-1} | z_{t-1}; \theta^{2}, \beta)
    & (t=21, \dots, 39), \\
    \mathrm{SBM} \, (x_{t}; \pi^{2}, \theta^{2}, K=3)
    & (t=40), \\
    \mathrm{LinkTrans} \, (x_{t-1} | z_{t-1}; \theta^{2}, \beta)
    & (t=41, \dots, 59), \\
    \mathrm{SBM} \, (x_{t}; \pi^{3}, \theta^{3}, K=4)
    & (t=60),  \\
    \mathrm{LinkTrans} \, (x_{t-1} | z_{t-1}; \theta^{3}, \beta) 
    & (t=61, \dots, 80),
    \end{cases}
\end{align}
where $\mathrm{SBM} \, (x; \pi, \theta, K)$ 
indicates that 
the model of SBM is $K$, 
with the mixture probability $\pi=(\pi_{1}, \dots, \pi_{K})$, 
and 
the link probability 
$\theta = \{ \theta_{k, \ell} \}_{k, \ell=1}^{K}$. 
Here, 
$\pi^{1}, \pi^{2} \in [0, 1]^{3}$, 
$\pi^{3} \in [0, 1]^{4}$, 
and 
$\theta^{1}, \theta^{2} \in [0, 1]^{3 \times 3}$, 
$\theta^{3} \in [0, 1]^{4 \times 4}$.  
We denote $\pi^{i} = (\pi^{i}_{1}, \dots, \pi^{i}_{K})$ \, 
$(i=1, 2, 3)$ 
and assume that 
$\pi_{i}^{1} \leq \dots \leq \pi_{i}^{K}$. 
Likewise, 
we denote $\theta^{i} = \{ \theta^{i}_{k, \ell} \}_{k, \ell=1}^{K}$ \, 
$(i=1, 2, 3)$. 
$\mathrm{LinkTrans}(x | z; \theta, \beta)$ 
means that 
some links are regenerated according to $\theta$ 
for each combination of blocks 
with probability $\beta$. 
In the following experiments, 
we set $\beta=0.02$.  
The hyperparameter of $\pi^{1}$ was set to $\alpha=1$ 
(the Dirichlet distribution), 
and those of $\theta^{1}$ were set to $a=b=1$ 
(the Beta distribution). 
$\theta^{2} = \{ \theta^{2}_{k, \ell} \}_{k, \ell=1}^{3}$ was set as follows: 
\begin{align}
\theta^{2}_{k, \ell} = 
  \begin{cases}
  \theta^{1}_{k, \ell} + u & (0 \leq \theta^{1}_{k, \ell} + u \leq 1), \\
  1-\epsilon & (\theta^{1}_{k, \ell} + u > 1),  \\
  \epsilon & (\theta^{1}_{k, \ell} + u < 0). 
  \end{cases}
\end{align}
Here, $u$ was drawn from the uniform distribution 
whose range is between $-0.1$ and $0.1$. 
$\theta^{3} = \{ \theta^{3}_{k, \ell} \}_{k, \ell=1}^{4}$ was set as follows:
\begin{align}
\theta^{3}_{k, \ell} &= 
  \begin{cases}
  \theta^{2}_{k, \ell} & (1 \leq k, \ell \leq 3), \\
  \theta^{3}_{k, 3} \sim \mathrm{Beta}(a, b) & (1 \leq k \leq 3, \ell = 4),  \\
  \theta^{3}_{3, \ell} \sim \mathrm{Beta}(a, b) & (k = 4, 1 \leq \ell \leq 4). 
  \end{cases}
\end{align}
It means that the link probabilities between the newly generated group 
$k=4$ and other blocks are the same as those between $k=3$ and 
other blocks. 
We set  
$\pi^{2} = (\pi_{1}^{2}, \pi_{2}^{2}, \pi_{3}^{2}) 
        =  \left(
             \pi_{1}^{1}, 
             \pi_{2}^{1} + (\pi_{3}^{1} - \pi_{2}^{1})/3, 
            \pi_{3}^{1} - (\pi_{3}^{1} - \pi_{2}^{1})/3
           \right)$ 
and 
$\pi^{3} = (\pi_{1}^{3}, \pi_{2}^{3}, \pi_{3}^{3}, \pi_{4}^{3}) 
        =  \left(
             \pi_{1}^{2}, \pi_{2}^{2}, 
             3\pi_{3}^{2}/4, 
             \pi_{3}^{2}/4
           \right)$. 
It means that the largest group $k=3$ is split into 
two blocks $k=3$, $4$ with the ratio of $3$ to $1$ at $t=60$. 
Fig.~\ref{fig:sample_X_abrupt} shows 
a sample sequence of $x_{t}$. 

\begin{figure}[b]
\centering
\includegraphics[width=\linewidth]{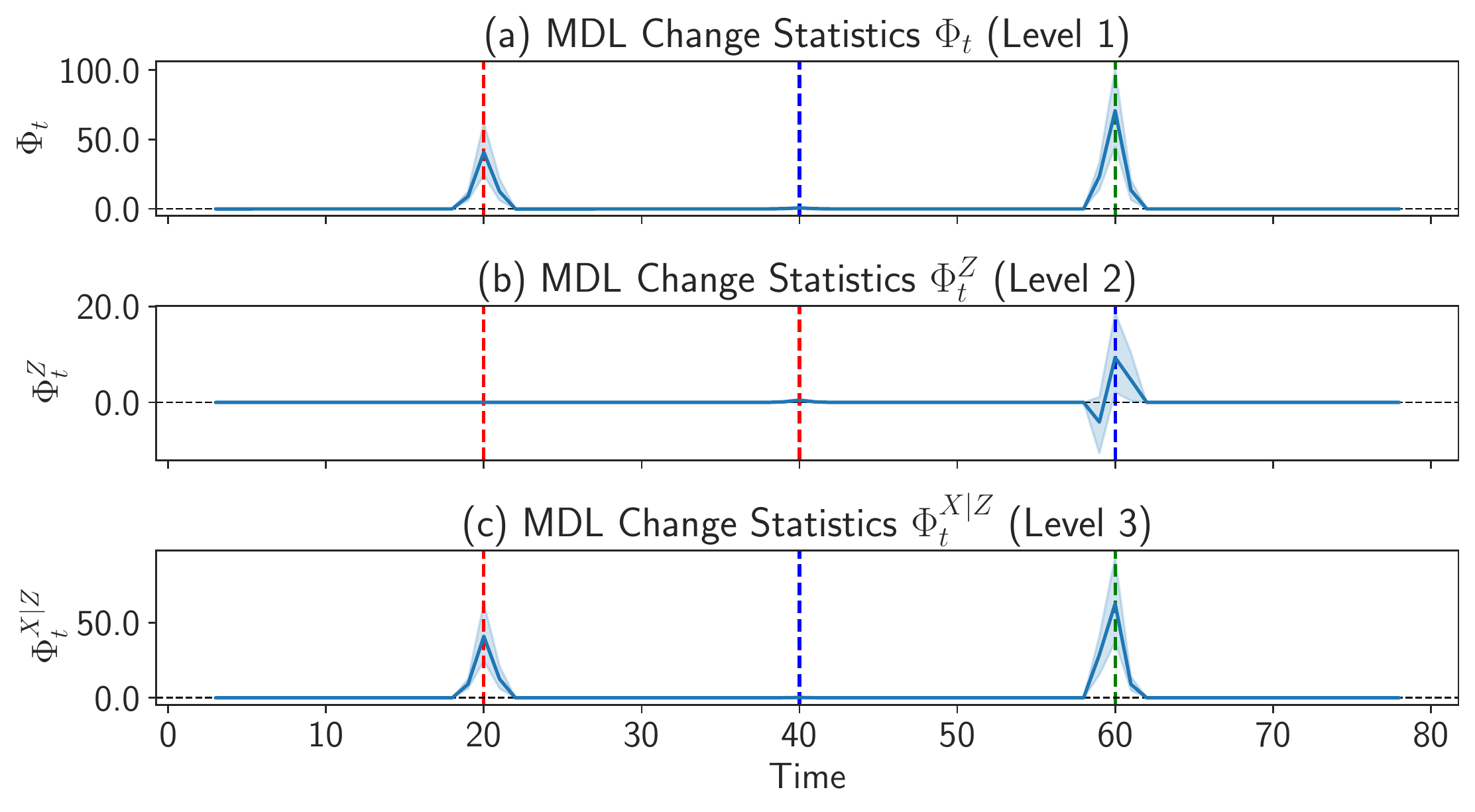}
\caption{
  MDL change statistics 
  for 20 datasets with abrupt changes 
  at $t=20$ (level 1), 
  $t=40$ (level 2), 
  and $t=60$ (level 3): 
  (a) $\Phi_{t}$, 
  (b) $\Phi^{Z}_{t}$, and 
  (c) $\Phi^{X|Z}_{t}$. 
  Window size $h=2$. 
}
\label{fig:plot_abrupt}
\end{figure}

We first estimated $\hat{z}_{t}$ at each $t$ 
with $x_{t}$ using the variational Bayes method \cite{Mariadassou2010}. 
As there might be a so-called \textit{label switching problem} (e.g., \cite{Redner1984,Jasra2005})
for $\{ \hat{z}_{t} \}_{t=1}$, 
we applied ECR algorithm \cite{Rodriguez2014,Papastamoulis2010} to reorder their blocks. 
We calculated the code-length for model $M$ as 
$L(M) = \log{2.865} + \log{M} + \log{\log{M}} + \dots$ in Eq.~\eqref{eq:def_MDL+DNML} \cite{Rissanen2012}, 
where the sum is taken for all the positive terms. 
We also calculated the code-length for models $M_{1}$ and $M_{2}$ as 
$L(M_{1}, M_{2}) = L(M_{1}) + L(M_{2} | M_{1})$. 
$L(M_{2} | M_{1})$ is code-length necessary for encoding $M_{2}$ given $M_{1}$, 
that is, 
\begin{align}
L(M_{2} | M_{1}) = 
  \begin{cases}
  -\log{ (1 - \alpha) } & (M_{1} = M_{2}), \\
  -\log{ \alpha/(K-1) } & (M_{1} \neq M_{2}), 
  \end{cases}
\end{align}
where $K$ denotes the maximum number of models, 
and $\alpha \mydef (N_{t} + 1/2) / (t+1)$ means the Krichevsky-Trofimov estimator~\cite{Krichevsky1981}. 
Here, $N_{t}$ is the number of model changes until $t-1$. 
We set $K=10$ 
and 
all the confidence parameters $\delta = \delta_{X|Z} = \delta_{Z} = 0.05$. 
We repeated the procedure for 20 times. 
Fig.~\ref{fig:plot_abrupt} shows 
the estimated number of blocks 
and the MDL change statistics 
$\Phi_{t}$, 
$\Phi^{Z}_{t}$, and $\Phi^{X|Z}_{t}$. 
We observe the following results from Fig.~\ref{fig:plot_abrupt}: 
\begin{itemize}
\item $\Phi_{t}$ changed at $t=20$, $40$ and $60$. 
      This means that abrupt changes occurred at level 1, 2, and 3, respectively. 
\item $\Phi^{Z}_{t}$ changed at $t=40$ and $60$. 
      This means that abrupt changes occurred at level 2 at $t=40$ 
      and at level 3 at $t=60$. 
\item $\Phi^{X|Z}_{t}$ changed at $t=20$ and $60$. 
      This means that abrupt changes occurred at level 1 at $t=20$ 
      and at level 3 at $t=60$. 
\end{itemize}

\subsubsection{Evaluation Metrics}
For evaluation, 
we investigated the trade-off 
between the detection delay and accuracy 
for a change point $t^{\ast}$ 
in terms of the benefit and the false alarm rate 
of a detected change point $\hat{t}$, 
which are defined as 
\begin{align}
\mathrm{benefit}
&=  \max \left( 1 - \frac{ |\hat{t} - t^{\ast}| }{T}, 0 \right) \quad ( \hat{t} > t^{\ast} ), 
\label{eq:def_benefit} \\
\mathrm{FAR} 
&= \frac{ 
     | \{ t : t^{\ast} - U < t < t^{\ast}, \phi_{t} > e \} |
   }{
     | \{ t : t^{\ast} - U < t < t^{\ast} \} |
   }, 
\label{eq:def_far}
\end{align}
where 
$(t^{\ast}, \phi_{t}, e) = (60, \Phi_{t}, \epsilon)$ (level 3), 
$(t^{\ast}, \phi_{t}, e) = (40, \Phi_{t}^{Z}, \epsilon_{Z})$ (level 2), 
and $(t^{\ast}, \phi_{t}, e) = (20, \Phi_{t}^{X|Z}, \epsilon_{X|Z})$ (level 1). 
$U$ denotes the period for which {\em overdetection} is not allowed. 
In this experiment, 
We set $T=5$ and $U=10$ in Eq.~\eqref{eq:def_benefit} and 
\eqref{eq:def_far}, respectively.

\subsubsection{Methods for Comparison}
We evaluated the performance of HCDL 
by comparing it with benchmark methods. 
We used the following three methods as benchmarks:
\begin{itemize}
\item {\em The conventional MDL change statistics-based method}~(NML). 
We used the MDL change statistics-based change detection, which employs the NML code-length in Eq.~\eqref{eq:NML} instead of Eq.~\eqref{eq:def_dnml}. We abbreviate this method as NML.
It can detect changes at level 3 only but can not decompose them in a hierachical way. 
Sakai and Yamanishi \cite{Sakai2013} 
proposed an efficient approximate formula for 
the generalized relation model including SBM, 
and we use it. 

\item {\em Tracking the Best Expert}~(TBE) \cite{Herbster1998}.  
Herbster and Warmuth developed the {\em fixed share algorithm}, abbreviated as FS. It was originally designed to make predictions by taking a weighted average over a number of experts.
In FS, the expert with the largest weight is the best expert.
We can think of FS as a model change detection 
algorithm by tracking the time-varying best expert.
We updated each weight $w_{t,k}^{s}$ 
for expert $k$ at time $t$ as
\begin{align}
w_{t, k}^{m} &= 
  w_{t, k}^{s} 
  e^{ 
      -\eta \left| 
             \mathrm{DNML}(x_{t}, z_{t}; K_{t}) -
             \mathrm{DNML}(x_{t}, \hat{z}_{t}; k) 
           \right| 
  }, \\
\mathrm{pool} &= \sum_{k=1}^{K} \alpha w_{t,i}^{m}, \\
w_{t+1, k}^{s} &= (1-\alpha) w_{t,k}^{m} + \frac{1}{K-1} (\mathrm{pool} - \alpha w_{t,i}^{m}), 
\end{align}
where $K_{t}$ means the true number of blocks at $t$. 
The ratio parameter $\alpha$ of TBE 
was fixed to $\alpha=0.2$. 

\item {\em DeltaCon} \cite{Koutra2016}. 
Koutra et al. proposed a graph similarity-based approach 
to detect change points in dynamic networks 
called {\em DeltaCon}. 
It calculates feature similarity of each consecutive snapshot pair of networks. 
We define the change score as $1-\mathrm{similarity}$. 

\end{itemize}

We determined the threshold parameters 
for benchmark methods. 
They were tuned to minimize the harmonic mean of 
the average benefit and $(1-$ the false alarm rate $)$ at $t=60$, 
at which a level 3 change occurred. 
For each procedure, 
we selected $\epsilon_{_\mathrm{NML}} \in \{ 0.1, 0.5, 1, 2, 5 \}$ 
for NML, 
$\epsilon_{_\mathrm{TBE}} \in \{ 0.2, 0.5, 0.8 \}$ 
and for TBE, 
$\epsilon_{_\mathrm{Delta}} \in \{ 0.02, 0.025, 0.03, 0.035, 0.04, 0.045, 0.05 \}$ 
for DeltaCon, 
respectively. 

Table~\ref{table:result_change_detection_for_dataset1}
lists the average benefits and false alarm rates for each algorithm 
and each task: model change detection (change in $K$) and 
parameter change detection (changes in $\theta$ and $\pi$). 
The experimental results demonstrated that 
our proposed HCDL algorithm was able to detect hierarchical changes 
and identify their levels, which could not be discriminated 
by the benchmark methods. 
Note that DeltaCon is competitive with HCDL except the benefit at $t=20$, 
but it can not inherently identify change levels. 

\begin{table*}[t]
\begin{center}
\begin{footnotesize}
\caption{Average benefits and FARs for each level of change
 for abrupt changes.}
\label{table:result_change_detection_for_dataset1}
\renewcommand{\arraystretch}{0.2}
{\tabcolsep=\tabcolsep
 \begin{tabular}{rrrrrrr}
 \toprule
 & \multicolumn{2}{c}{Level~3 ($t=60$)} &
   \multicolumn{2}{c}{Level~2 ($t=40$)} &
   \multicolumn{2}{c}{Level~1 ($t=20$)} \\
 & \multicolumn{1}{c}{benefit} & \multicolumn{1}{c}{FAR} &
   \multicolumn{1}{c}{benefit} & \multicolumn{1}{c}{FAR} &
   \multicolumn{1}{c}{benefit} & \multicolumn{1}{c}{FAR}  \\
 \midrule
 \multicolumn{1}{c}{HCDL} ($h=1$) &
 $\mathbf{1.00 \pm 0.00}$ & 
 $\mathbf{0.00 \pm 0.00}$ &
 $\mathbf{1.00 \pm 0.00}$ & 
 $\mathbf{0.00 \pm 0.00}$ &
 $0.97 \pm 0.03$ & 
 $\mathbf{0.00 \pm 0.00}$ \\
 \multicolumn{1}{c}{HCDL} ($h=2$) &
 $\mathbf{1.00 \pm 0.00}$ & 
 $\mathbf{0.00 \pm 0.00}$ &
 $\mathbf{1.00 \pm 0.00}$ & 
 $\mathbf{0.00 \pm 0.00}$ &
 $\mathbf{1.00 \pm 0.00}$ & 
 $\mathbf{0.00 \pm 0.00}$ \\
 \multicolumn{1}{c}{HCDL} ($h=3$) &
 $\mathbf{1.00 \pm 0.00}$ & 
 $\mathbf{0.00 \pm 0.00}$ &
 $\mathbf{1.00 \pm 0.00}$ & 
 $\mathbf{0.00 \pm 0.00}$ &
 $\mathbf{1.00 \pm 0.00}$ & 
 $\mathbf{0.00 \pm 0.00}$ \\
 \midrule
 \multicolumn{1}{c}{NML} ($h=1$) &
 $\mathbf{1.00 \pm 0.00}$ & 
 $0.03 \pm 0.02$ & 
 $0.61 \pm 0.04$ &
 $\mathbf{0.00 \pm 0.00}$ & 
 $0.96 \pm 0.02$ &
 $\mathbf{0.00 \pm 0.00}$ \\
 \multicolumn{1}{c}{NML} ($h=2$) &
 $\mathbf{1.00 \pm 0.00}$ & 
 $0.04 \pm 0.03$ &
 $0.63 \pm 0.07$ &  
 $0.03 \pm 0.05$ &
 $\mathbf{1.00 \pm 0.00}$ & 
 $\mathbf{0.00 \pm 0.00}$ \\
 \multicolumn{1}{c}{NML} ($h=3$) &
 $\mathbf{1.00 \pm 0.00}$ & 
 $\mathbf{0.00 \pm 0.00}$ &
 $0.64 \pm 0.08$ &  
 $0.03 \pm 0.05$ &
 $\mathbf{1.00 \pm 0.00}$ &  
 $\mathbf{0.00 \pm 0.00}$ \\
 \midrule
 \multicolumn{1}{c}{TBE} &
 $0.44 \pm 0.37$ & 
 $\mathbf{0.00 \pm 0.00}$ &
 $0.00 \pm 0.00$ & 
 $\mathbf{0.00 \pm 0.00}$ &
 $0.00 \pm 0.00$ & 
 $\mathbf{0.00 \pm 0.00}$ \\
 \midrule
 \multicolumn{1}{c}{DeltaCon} &
 $\mathbf{1.00 \pm 0.00}$ & 
 $\mathbf{0.00 \pm 0.00}$ &
 $\mathbf{1.00 \pm 0.00}$ & 
 $\mathbf{0.00 \pm 0.00}$ &
 $0.85 \pm 0.37$ & 
 $\mathbf{0.00 \pm 0.00}$ \\
 \bottomrule
\end{tabular}
}
\end{footnotesize}
\end{center}
\end{table*}

\subsection{Synthetic Dataset 2 (Gradual Change)}
\label{subsection:synthetic_dataset_2}

We generated a dataset
in which each level of change occurs gradually.

\begin{figure}[tb]
\centering
\includegraphics[width=\linewidth]{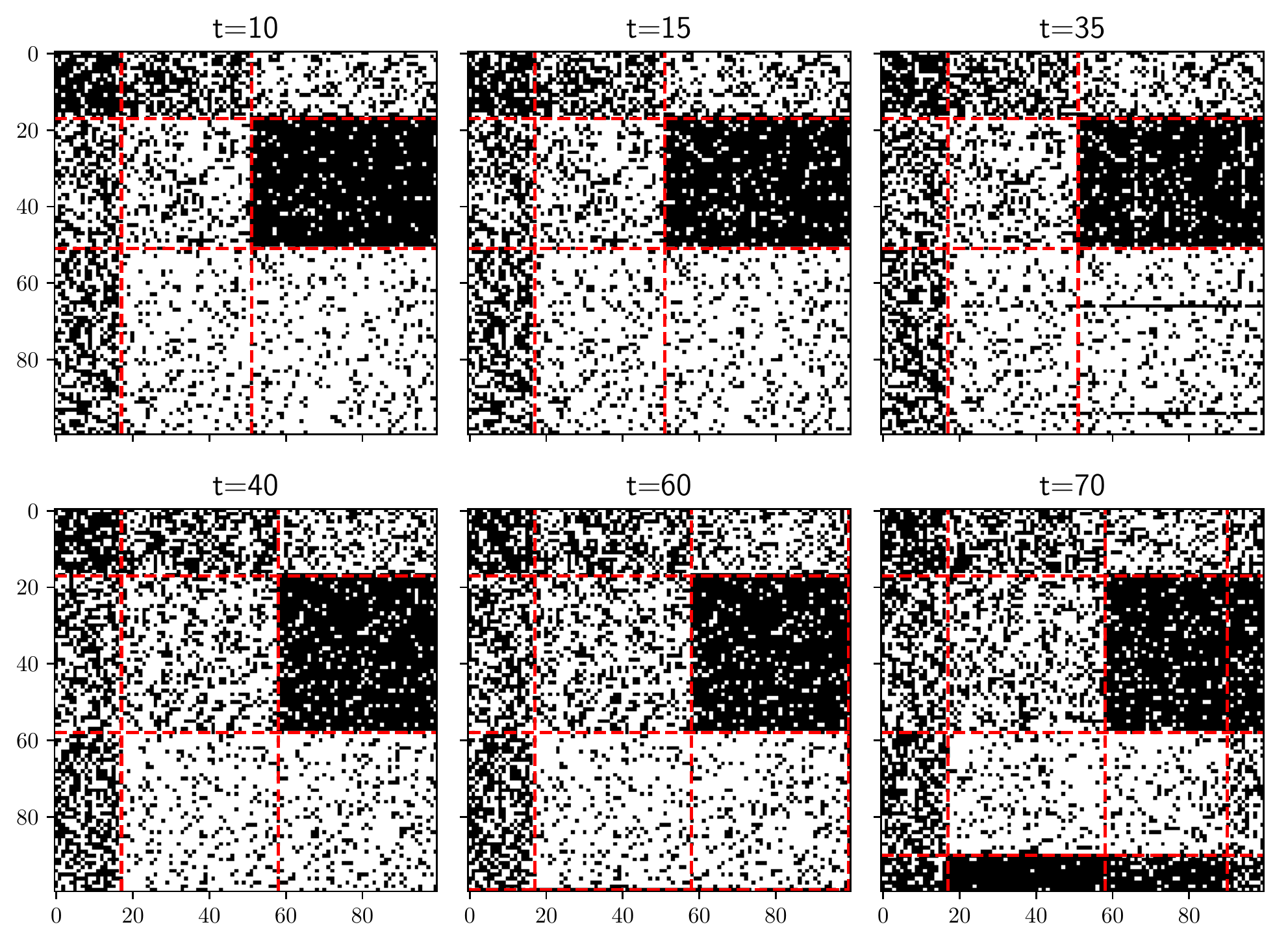}
\caption{ A sample sequence of links $x_{t}$ for gradual change.
 Link probability gradually changes between $t=10$ and $15$.
 Mixture probability gradually changes between $t=35$ and $40$.
 At $t=60$, model changes from $K=3$ to $K=4$,
 and mixture probability of newly generated group
 gradually increases from $t=60$ to $70$.
}
\label{fig:sample_X_gradual}
\end{figure}

For $t=1, \dots, 90$,
we generated links $x_{t}$ between nodes
as follows:
\begin{align}
x_{t} \sim
  \renewcommand{\arraystretch}{0.7}
  \begin{cases}
    \mathrm{SBM} \, (x_{t}; \pi^{1}, \theta^{1}, K=3)
    & (t=1), \\
    \mathrm{LinkTrans} (x_{t} | z_{t-1}; \pi^{1}, \theta^{1}, \beta)
    & (t=2, \dots, 9), \\
    \mathrm{SBM} \, (x_{t}; \pi^{1}, \theta^{1,2}(t), K=3)
    & (t=10, \dots, 15), \\
    \mathrm{LinkTrans} (x_{t} | z_{t-1}; \pi^{1}, \theta^{2}, \beta)
    & (t=16, \dots, 34), \\
    \mathrm{SBM} \, (x_{t}; \pi^{1,2}(t), \theta^{2}, K=3)
    & (t=35, \dots, 40), \\
    \mathrm{LinkTrans} (x_{t} | z_{t-1}; \pi^{2}, \theta^{2}, \beta)
    & (t=41, \dots, 59), \\
    \mathrm{SBM} \, (x_{t}; \pi^{2,3}(t), \theta^{3}, K=4)
    & (t=60, \dots, 70), \\
    \mathrm{LinkTrans} (x_{t} | z_{t-1}; \pi^{3}, \theta^{2}, \beta)
    & (t=71, \dots, 90),
  \end{cases}
  \nonumber
  \label{eq:data_generation_gradual}
\end{align}
where $\mathrm{SBM}$ and $\mathrm{LinkTrans}$
are the same as in the abrupt change case in
Section~\ref{subsection:synthetic_dataset_1}.
The link probability $\theta$  gradually changes
between $t=10$ and $t=15$.
Then,
the mixture probability $\pi$ gradually changes
between $t=35$ and $t=40$.
Finally,
the model changes from $K=3$ to $K=4$ at $t=60$,
and then
the mixture probability of the newly generated blocks 
gradually increases from $t=60$ to $t=70$.
Fig.~\ref{fig:sample_X_gradual} shows
a sample sequence of $x_{t}$.

The hyperparameter of $\pi^{1}$ was set to $\alpha=1$
(the Dirichlet distribution),
and those of $\theta^{1}$ were set to $a=b=1$
(the Beta distribution).
We set 
$\pi^{2} = 
   (\pi_{1}^{1}, 
   (\pi_{2}^{1} + \pi_{3}^{1})/2, 
   (\pi_{2}^{1} + \pi_{3}^{1})/2)$ 
and 
$\pi^{3} = 
   (\pi_{1}^{3}, 
    \pi_{2}^{3}, 
    \pi_{3}^{3}, 
    \pi_{4}^{3})
 = (\pi_{1}^{2}, 
    \pi_{2}^{2}, 
     3\pi_{3}^{2}/4, 
     \pi_{3}^{2}/4)$. 
It means that the largest group $k=3$ is split into two blocks 
with the ratio of $3$ to $1$ between $t=60$ and $t=70$.
$\theta^{2}$ $(k, \ell = 1, \dots, 4)$ was set as follows:
$\theta^{2}_{k, \ell} =
  \theta^{1}_{k, \ell} + u$ $(0 \leq \theta^{1}_{k, \ell} + u \leq 1)$,
  $1 - \epsilon$ $(\theta^{1}_{k, \ell} + u > 1)$,
  and $\epsilon$ $(\theta^{1}_{k, \ell} + u < 0)$.
Here,
$u$ was drawn from the uniform distribution whose range is between $-0.1$ and $0.1$,
and
$\epsilon$ was set to $\epsilon=10^{-6}$.
$\theta^{3}$ $(k, \ell = 1, \dots, 4)$ was set as follows:
$\theta^{3}_{k, \ell} =
  \theta^{2}_{k, \ell}$ $(1 \leq k, \ell \leq 3)$,
$\theta^{3}_{k, 4} \sim \mathrm{Beta}(a, b)$ $(1 \leq k \leq 4)$,
and
$\theta^{3}_{4, \ell} \sim \mathrm{Beta}(a, b)$ $(1 \leq \ell \leq 3)$.

From $t=10$ to $t=15$,
$\theta^{1,2}(t)$ is defined as
$\theta^{1,2}(t) = \theta^{1} + (t-10)(\theta^{2} - \theta^{1})/5$,
which means that $\theta$ started to change at $t=10$
and finally reaches $\theta^{2}$ at $t=15$.
Likewise,
from $t=35$ to $t=40$,
$\pi^{1, 2}(t)$ was defined as
$\pi^{1,2}(t)
 = ( \pi^{1}_{1},
     \pi^{1}_{2} + (t-35)(\pi^{1}_{3} - \pi^{1}_{2})/10,
     \pi^{1}_{3} - (t-35)(\pi^{1}_{3} - \pi^{1}_{2})/10
   )
$,
which means that
the third component of $\pi$ gradually decreases.
Finally, the model changes at $t=60$
from $K=3$ to $K=4$.
Then, the mixture probability $\pi$ gradually changes
between $t=60$ and $70$ as
$\pi^{2,3}(t)
 = (
     \pi^{2}_{1},
     \pi^{2}_{2},
     \pi^{2}_{3} - (t-60)\pi^{2}_{3}/40,
     (t-60) \pi^{2}_{3}/40
   )
$,
which means that
the third component of $\pi$ decreases gradually,
and finally the ratio of the third and the newly fourth components
reaches $3$ to $1$.

\begin{figure}[tb]
\centering
\includegraphics[width=\linewidth]{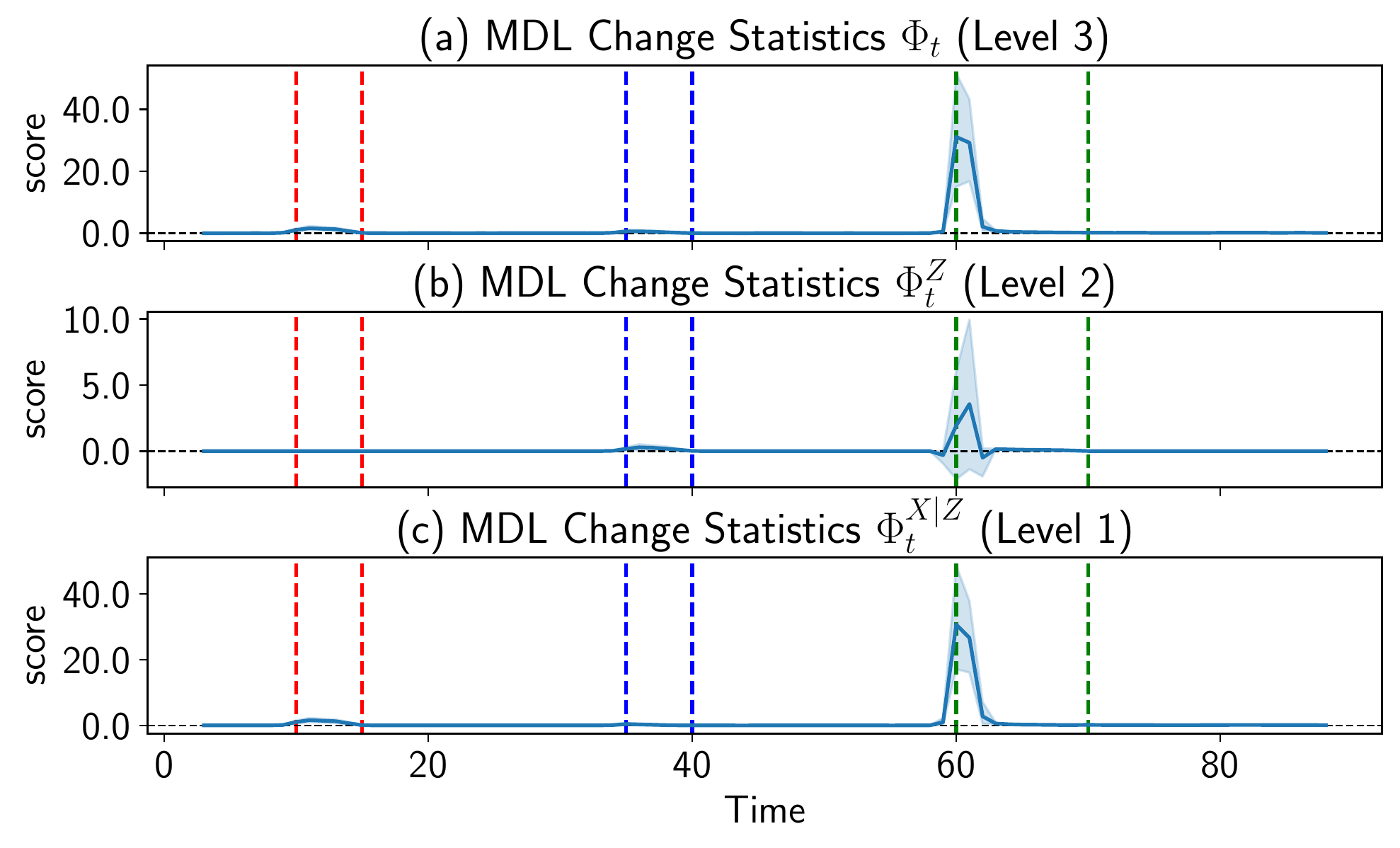}
\caption{ MDL change statistics
          for 20 datasets with gradual changes
          from $t=10$ to $15$ (level 1),
          from $t=35$ to $40$ (level 2),
          and from $t=60$ to $70$ (level 3):
          (a) $\Phi_{t}$,
          (b) $\Phi_{t}^{Z}$, and
          (c) $\Phi_{t}^{X|Z}$.
          Window size $h=2$.
}
\label{fig:plot_gradual}
\end{figure}

We repeated the procedure for 20 times.
We set
the maximum number of models $K=10$
and
all the confidence parameters
$\delta = \delta_{X|Z} = \delta_{Z} = 0.05$.
Fig.~\ref{fig:plot_gradual} shows
the estimated number of blocks and
the MDL change statistics $\Phi_{t}$,
$\Phi_{t}^{Z}$,
and
$\Phi_{t}^{X|Z}$.
We observe the following results from Fig.~\ref{fig:plot_gradual}:
\begin{itemize}
\item $\Phi_{t}$ started to change gradually at $t=10$, $35$, and $60$.
\item $\Phi_{t}^{X|Z}$ started to change gradually at $t=10$ and $60$.
      This means that changes at level 1 were found.
\item $\Phi_{t}^{Z}$ started to change gradually at $t=35$ and $60$.
      This means that changes at level 2 were found.
\end{itemize}

\subsubsection{Evaluation Metrics}
For evaluation,
we investigated the trade-off
between the benefits in Eq.~\eqref{eq:def_benefit}
and false alarm rate in Eq.~\eqref{eq:def_far}
for a change point $t^{\ast}$.
Here, $(t^{\ast}, \phi_{t}, e) = (60, \Phi_{t}, \epsilon)$ (level 3),
      $(t^{\ast}, \phi_{t}, e) = (35, \Phi_{t}^{Z}, \epsilon_{Z})$ (level 2),
and  $(t^{\ast}, \phi_{t}, e) = (10, \Phi_{t}^{X|Z}, \epsilon_{X|Z})$ (level 1). 
We set $T=5$ and $U=10$ in Eq.~\eqref{eq:def_benefit}
and \eqref{eq:def_far}, respectively.

\subsubsection{Methods for Comparison}
We evaluated the performance of HCDL
by comparing it with NML, TBE, and DeltaCon 
as in the abrupt change case.
We selected the threshold parameters for each method
by tuning them among the same values as in the abrupt change case.

Table~\ref{table:result_change_detection_for_dataset2} lists
the benefits and false alarm rates of
each algorithm.
This shows the effectiveness of HCDL for a dataset
in which the models and parameters are changing gradually.
Note that NML, TBE, and DeltaCon 
can not identify the levels of changes;
hence, their alarms are mixed for different levels.
We evaluated these alarms for each of the different levels.
Here again,
we observe that the
HCDL algorithm detected hierarchical changes and
identified their levels,
which could not be discriminated by the benchmark methods.

\begin{table*}[t]
\begin{center}
\begin{footnotesize}
\caption{Average benefits and FARs for each level of change
 for gradual changes.}
\label{table:result_change_detection_for_dataset2}
\renewcommand{\arraystretch}{0.2}
{\tabcolsep=\tabcolsep
 \begin{tabular}{rrrrrrr}
 \toprule
 & \multicolumn{2}{c}{Level~3 ($t=60-70$)} &
   \multicolumn{2}{c}{Level~2 ($t=35-40$)} &
   \multicolumn{2}{c}{Level~1 ($t=10-15$)} \\
 & \multicolumn{1}{c}{benefit} & \multicolumn{1}{c}{FAR} &
   \multicolumn{1}{c}{benefit} & \multicolumn{1}{c}{FAR} &
   \multicolumn{1}{c}{benefit} & \multicolumn{1}{c}{FAR}  \\
 \midrule
 \multicolumn{1}{c}{HCDL} ($h=1$) &
 $0.97 \pm 0.09$ & $\mathbf{0.00 \pm 0.00}$ &
 $0.96 \pm 0.02$ & $\mathbf{0.00 \pm 0.00}$ &
 $0.95 \pm 0.04$ & $\mathbf{0.00 \pm 0.00}$ \\
 \multicolumn{1}{c}{HCDL} ($h=2$) &
 $0.98 \pm 0.08$ & $\mathbf{0.00 \pm 0.00}$ &
 $0.97 \pm 0.03$ & $\mathbf{0.00 \pm 0.00}$ &
 $0.98 \pm 0.05$ & $\mathbf{0.00 \pm 0.00}$ \\
 \multicolumn{1}{c}{HCDL} ($h=3$) &
 $\mathbf{1.00 \pm 0.00}$ & $\mathbf{0.00 \pm 0.00}$ &
 $\mathbf{0.97 \pm 0.02}$ & $\mathbf{0.00 \pm 0.00}$ &
 $\mathbf{1.00 \pm 0.00}$ & $\mathbf{0.00 \pm 0.00}$ \\
 \midrule
 \multicolumn{1}{c}{NML} ($h=1$) &
 $0.89 \pm 0.10$ &  $\mathbf{0.00 \pm 0.00}$ &
 $0.61 \pm 0.04$ &  $\mathbf{0.00 \pm 0.00}$ &
 $0.69 \pm 0.10$ &  $0.04 \pm 0.05$ \\
 \multicolumn{1}{c}{NML} ($h=2$) &
 $0.90 \pm 0.10$ &  $\mathbf{0.00 \pm 0.00}$ &
 $0.62 \pm 0.06$ &  $\mathbf{0.00 \pm 0.00}$ &
 $0.70 \pm 0.10$ &  $0.04 \pm 0.05$ \\
 \multicolumn{1}{c}{NML} ($h=3$) &
 $0.91 \pm 0.09$ &  $\mathbf{0.00 \pm 0.00}$ &
 $0.63 \pm 0.07$ &  $0.01 \pm 0.02$ &
 $0.71 \pm 0.10$ &  $0.06 \pm 0.05$ \\
 \midrule
 \multicolumn{1}{c}{TBE} &
 $0.04 \pm 0.10$ & 
 $\mathbf{0.00 \pm 0.00}$ &
 $0.00 \pm 0.00$ & 
 $0.02 \pm 0.09$ &
 $0.00 \pm 0.00$ & 
 $0.10 \pm 0.30$ \\
 \midrule
 \multicolumn{1}{c}{DeltaCon} &
 $0.89 \pm 0.31$ & 
 $\mathbf{0.00 \pm 0.00}$ &
 $0.15 \pm 0.37$ & 
 $0.01 \pm 0.02$ &
 $0.64 \pm 0.48$ & 
 $0.01 \pm 0.03$ \\
 \bottomrule
\end{tabular}
}
\end{footnotesize}
\end{center}
\end{table*}

\subsection{Population Movement Dataset}

We demonstrate the effectiveness of HCDL on the population movement dataset
\footnote{\url{https://www.e-stat.go.jp/en/stat-search/database?page=1&toukei=00200523&tstat=000000070001}}. 
This dataset is presented by Ministry of Internal Affairs and Communications, 
Statistics Bureau, 
Director-General for Policy Planning \& Statistical Research and Training Institute, Japan. 
It records the amount of the population that moved from a prefecture 
to another prefecture per month from April 2005 to March 2014. 
The total numbers of time points and prefectures are 108 and 47, respectively. 
We conducted HCDL on the dataset. 
In this experiment, 
we assumed that
each number of population movement in each month between prefectures
was drawn from the Poisson distribution.
We set the window size $h=3$ for HCDL.

\begin{figure}
\centering
\includegraphics[width=\linewidth, angle=0.5]{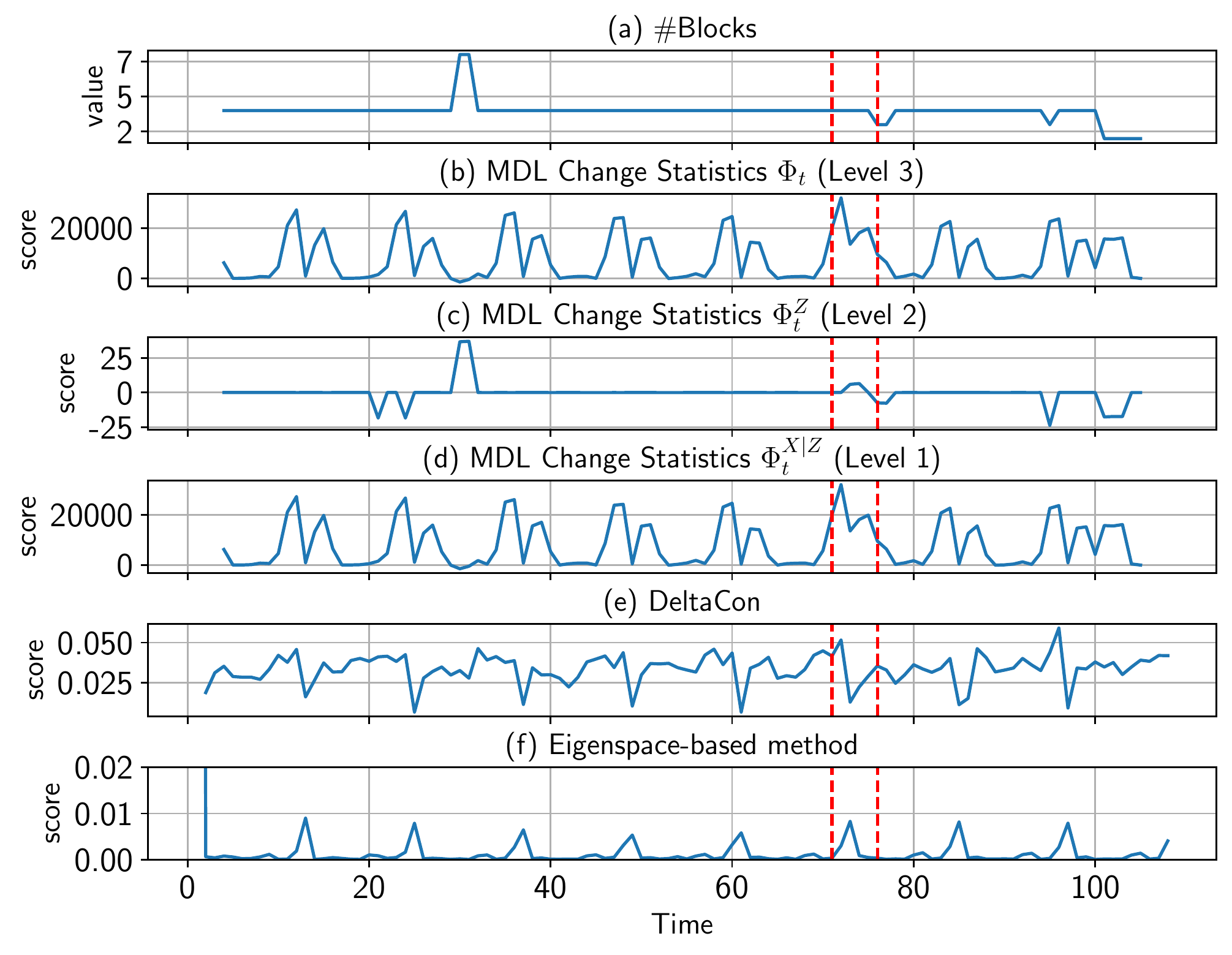}
\caption{Estimated number of blocks and change scores for Population Movement dataset.
 (a) Estimated number of blocks.
 (b) MDL change statistics $\Phi_{t}$.
 (c) $\Phi^{Z}_{t}$.
 (d) $\Phi^{X|Z}_{t}$.
 (e) change scores of DeltaCon \cite{Koutra2016}.
 (f) change scores of Eigenspace-based method \cite{Ide2004}.
 Dashed red line indicates a period between March and August, 2011.
}
\label{fig:estimated_number_of_blocks_for_popmove_dataset}
\end{figure}

Fig.~\ref{fig:estimated_number_of_blocks_for_popmove_dataset}
shows that HCDL detected the changes between $t=72$ and $t=76$,
that is, from March to August in 2011.
During the period
the MDL changes statistics $\Phi_{t}$
increased more than other years.
These changes corresponded to the population movement caused
by the Great East Japan Earthquake in March 2011,
the Japanese government's announcement that the radiological dosage levels
greatly increased in many regions in Japan in May 2011,
and the announcement of
population movement during June--August 2011
by Japanese government.
It is particularly noticeable that $\Phi_{t}^{Z}$ increased
in May and June 2011.
It is hard to extract such information with the rival algorithms,
DeltaCon \cite{Koutra2016} and
Eigenspace-based method \cite{Ide2004},
as shown in Fig.~\ref{fig:estimated_number_of_blocks_for_popmove_dataset} (d) and (e), 
because these algorithms can not decompose the change scores into each layer.

\begin{figure}
\centering
\includegraphics[width=\linewidth]{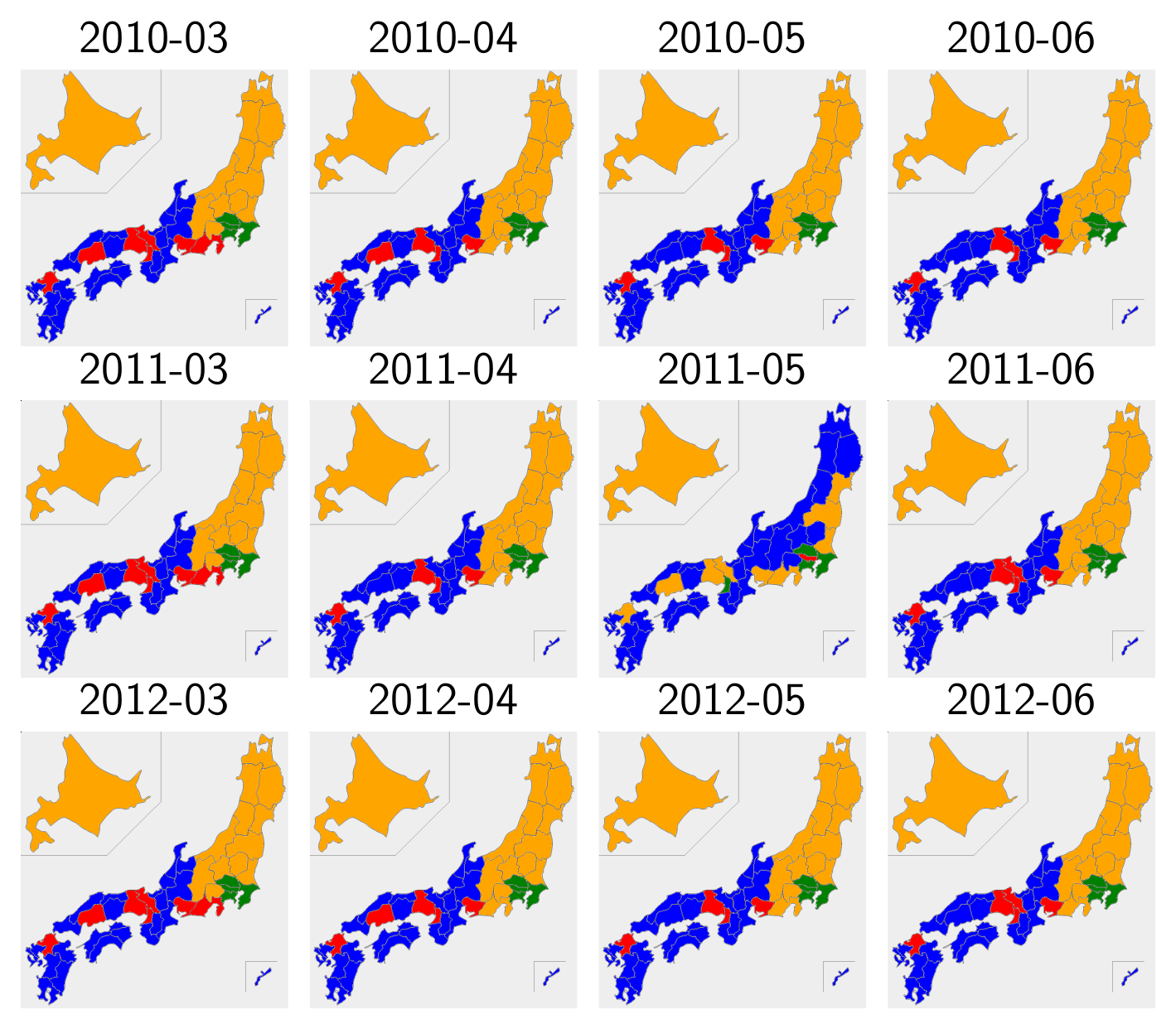}
\caption{Estimated blocks for the Population Movement dataset between January and August in 2010, 2011, and 2012.
The estimated blocks were different in May 2011
from these in other months in the same year or other years.
}
\label{fig:plot_pref_groups_2011_2013_01_08}
\end{figure}

We then investigated why the changes scores $\Phi_{t}$ were
higher in 2011 than those in other years.
Fig.~\ref{fig:plot_pref_groups_2011_2013_01_08} shows
the estimated blocks of each prefecture
for the Population Movement dataset
between January and August in 2010, 2011, and 2012.
Each color represents each group for all the prefectures.
We observe from Fig.~\ref{fig:plot_pref_groups_2011_2013_01_08}
that the estimated blocks were different in May 2011,
from these in other months in the same year or other years.
We inferred that this led to the rise of $\Phi_{t}^{Z}$ in May and June, 2011.
This corresponded to a change at level 2.

\begin{figure}
\centering
\includegraphics[width=\linewidth]{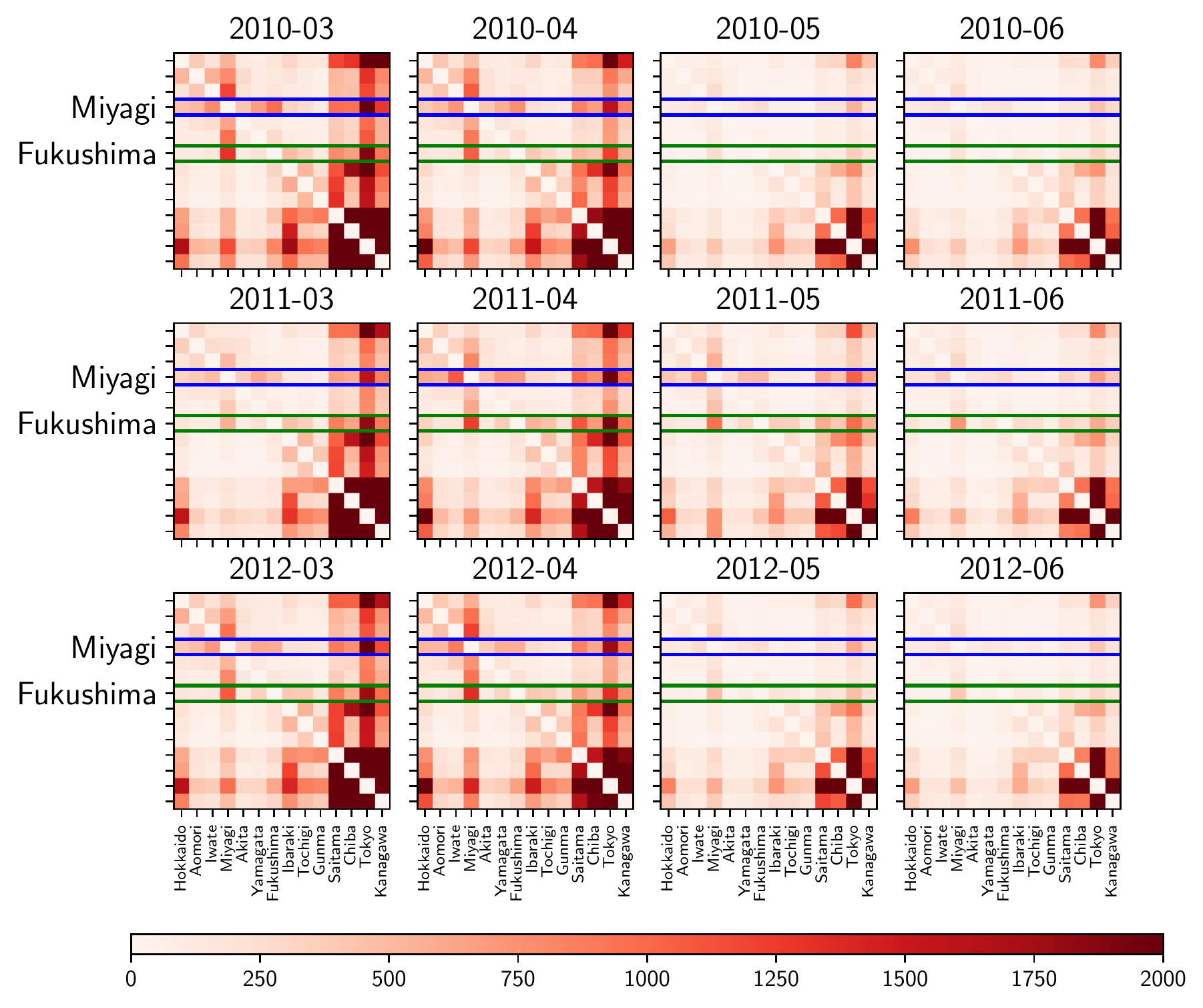}
\caption{Heatmaps of population moves among 14 prefectures between March and June in 2010, 2011, and 2012.
The vertical and horizontal axes represent the origin and destination prefectures, respectively.
The numbers are truncated at 2000.
More people moved from Miyagi prefecture and Fukushima prefecture
than other years,
in April and May 2011.
}
\label{fig:heatmap_popmove_hokkaido_kanagawa_2010_2012_03_06}
\end{figure}

We further investigated what change occurred at the population movement,
that is,
at level 1 during the period.
Fig.~\ref{fig:estimated_number_of_blocks_for_popmove_dataset} (d) shows
that $\Phi^{X|Z}_{t}$ increased more than these during the same months
in other years.
Fig.~\ref{fig:heatmap_popmove_hokkaido_kanagawa_2010_2012_03_06} shows
the heatmaps of the numbers of population move among 14 prefectures
between March and June in 2010, 2011, and 2012.
Note that the numbers of the population movement are
truncated at 2000 to easily compare the result.
We observe from Fig.~\ref{fig:heatmap_popmove_hokkaido_kanagawa_2010_2012_03_06} that
more people moved in April and May 2011 than other years,
from Miyagi prefecture and Fukushima prefecture
to the other prefectures.
These two prefectures were hardest hit by the disaster.
This corresponded to the change at level 1.

In summary,
we are able to interpret the events behind the changes
in terms of severity levels in the HCDL framework.

\subsection{ Enron Dataset }

We demonstrate the effectiveness of 
HCDL on a dynamic social network constructed 
from the Enron corpus \cite{Priebe2005} 
\footnote{\url{http://www.cs.cmu.edu/~enron/}}, 
which consists of about 0.5 million email messages. 
In this experiment, 
we extracted e-mails among 151 employees 
from April 1999 to August 2002.  
We preprocessed the corpus 
and obtained a connection matrix for each week 
by assigning 1 if employee $i$ sent at least one email 
to $j$ during the week, 
and 0 otherwise. 
We made no distinction between emails 
sent ``To'' and ``Cc.'' 

\begin{figure}[tb]
\begin{center}
\includegraphics[width=\linewidth]{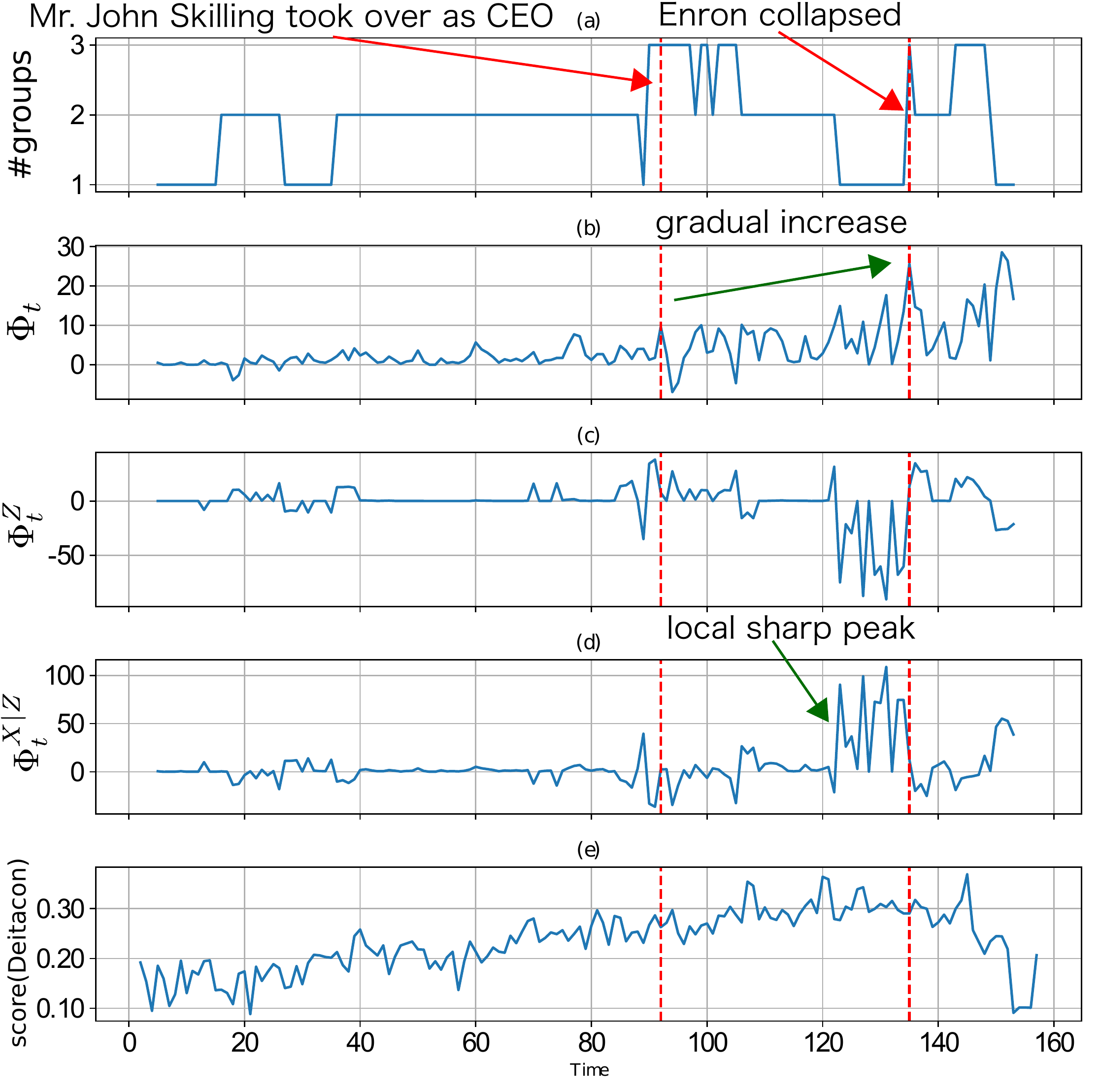}
\caption{Estimated blocks and change scores of the Enron dataset.
  (a) Estimated number of blocks. 
  (b) MDL change statistics $\Phi_{t}$. 
  (c) $\Phi^{Z}_{t}$. 
  (d) $\Phi^{X|Z}_{t}$. 
  (e) change scores of DeltaCon \cite{Koutra2016}. 
}
\label{fig:estimated_groups_and_change_scores_of_the_enron_dataset}
\end{center}
\end{figure}

Fig.~\ref{fig:estimated_groups_and_change_scores_of_the_enron_dataset} lists 
the estimated change scores for the Enron dataset. 
The horizontal axis represents the week index, 
while the vertical axis represents the change scores 
$\Phi_{t}$, $\Phi^{Z}_{t}$, and $\Phi^{X|Z}_{t}$. 
We set the window parameter $h=4$. 
We observe from Fig.~\ref{fig:estimated_groups_and_change_scores_of_the_enron_dataset} that, 
from $t=80$ to $t=90$, $\Phi^{Z}_{t}$ and $\Phi^{X|Z}_{t}$ showed local sharp peaks, 
 while $\Phi_{t}$ was relatively mild. 
Actually, 
Mr. Jeffrey Skilling took over as CEO in February 2001. 
It shows that this event corresponded to changes at levels 1 and 2.
Around $t=110$ to $150$, all of $\Phi_{t}$, $\Phi^{X|Z}_{t}$ and $\Phi^{Z}_{t}$ had sharp peaks at several points.
Actually, at $t=135$, 
Enron collapsed in December 2001, and a number of related events occurred around this time.
It shows that this event corresponded to changes at level 3.

\begin{figure}[tb]
\begin{center}
\includegraphics[width=\linewidth]{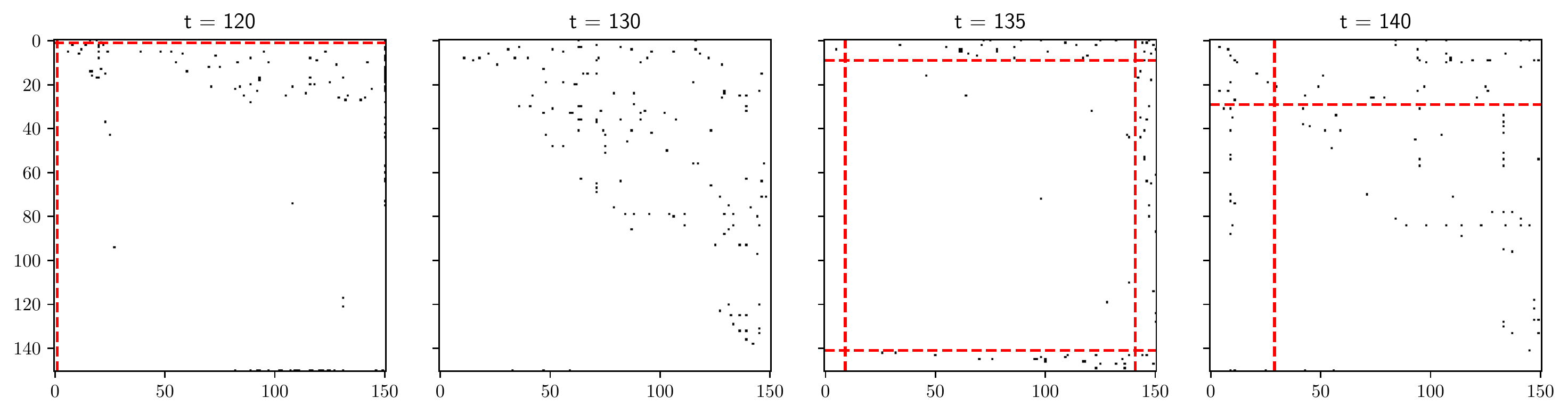}
\caption{ Visualization of messages between employees and the estimated blocks at $t=120$, $130$, $135$, and $140$.}
\label{fig:visualization_of_messages_between_employees_and_the_estimated_groups_for_the_enron_dataset}
\end{center}
\end{figure}

We see from Fig.~\ref{fig:estimated_groups_and_change_scores_of_the_enron_dataset} 
that before and after Enron collapsed at $t=135$, 
the number of blocks rapidly changed. 
We could see the messages between employees changed dynamically. 
We also see from Fig.~\ref{fig:visualization_of_messages_between_employees_and_the_estimated_groups_for_the_enron_dataset}
that the number of blocks rapidly changed 
before and after Enron collapsed at $t=135$. 

We could see the number of messages or relation between employees 
changed dynamically 
and it affected the number of group estimated with HCDL 
and the structures of blocks. 
It owes to the decomposed nature of DNML. 

In this way, we are able to interpret the events behind the changes in terms of severity levels in the HCDL framework.

%% file: 7_Conclusion.tex
\section{Conclusion}
This study established HCDL 
to detect hierarchical changes in latent variable models. 
HCDL aims at detecting and explaining those changes.  
The idea 
is to employ an information-theoretic framework 
for change detection 
in which 
the MDL change statistics is used 
as a measure of the degree of change, 
in combination with the DNML code-length. 
We presented a rationale for 
reliable alarms of changes 
on the theory for hypothesis testing. 
With synthetic and real 
datasets, 
we demonstrated that 
HCDL is highly effective.


%% file: 8_Acknowledgement.tex
\section*{Acknowledgement}
This work was partially supported by JST KAKENHI 191400000190 and JST-AIP JPMJCR19U4.